\documentclass[10pt,twocolumn,letterpaper]{article}

\usepackage[pagenumbers]{cvpr} 

\usepackage{siunitx, booktabs}
\usepackage{adjustbox}
\usepackage[table]{xcolor}

\definecolor{cvprblue}{rgb}{0.21,0.49,0.74}
\usepackage[pagebackref,breaklinks,colorlinks,citecolor=cvprblue]{hyperref}

\definecolor{molmocolor}{RGB}{240, 82, 156}
\definecolor{tablegray}{RGB}{223, 242, 252}

\usepackage{array, graphicx, amsmath, amssymb, caption, subcaption, multirow, overpic, textpos}

\newlength\savewidth\newcommand\shline{\noalign{\global\savewidth\arrayrulewidth
  \global\arrayrulewidth 1pt}\hline\noalign{\global\arrayrulewidth\savewidth}}
\newcommand{\tablestyle}[2]{\setlength{\tabcolsep}{#1}\renewcommand{\arraystretch}{#2}\centering\footnotesize}
\renewcommand{\paragraph}[1]{\vspace{0.5mm}\noindent\textbf{#1}}

\newcolumntype{x}[1]{>{\centering\arraybackslash}p{#1pt}}
\newcolumntype{y}[1]{>{\raggedright\arraybackslash}p{#1pt}}
\newcolumntype{z}[1]{>{\raggedleft\arraybackslash}p{#1pt}}

\newcommand{\app}{\raise.17ex\hbox{$\scriptstyle\sim$}}

\newcommand{\x}{{\times}}
\definecolor{deemph}{gray}{0.6}

\definecolor{baselinecolor}{gray}{.9}
\newcommand{\baseline}[1]{\cellcolor{baselinecolor}{#1}}

\def\paperID{****} 
\def\confName{CVPR}
\def\confYear{2025}

\title{\vspace{-2mm}\textit{Molmo and PixMo}:\\Open Weights and Open Data \\for State-of-the-Art Vision-Language Models\vspace{-4.5mm}}

\author{
Matt Deitke$^{*\dagger\psi}$ \quad  Christopher Clark$^{*\dagger}$ \quad  Sangho Lee$^{\dagger}$ \quad  Rohun Tripathi$^{\dagger}$ \quad  Yue Yang$^{\dagger}$ \\
Jae Sung Park$^{\psi}$ \quad Mohammadreza Salehi$^{\psi}$ \quad  Niklas Muennighoff$^{\dagger}$ \quad  Kyle Lo$^{\dagger}$ \quad  Luca Soldaini$^{\dagger}$ \\ 
Jiasen Lu$^{\dagger}$ \quad Taira Anderson$^{\dagger}$ \quad Erin Bransom$^{\dagger}$ \quad Kiana Ehsani$^{\dagger}$ \quad  Huong Ngo$^{\dagger}$ \\
YenSung Chen$^{\dagger}$ \quad Ajay Patel$^{\dagger}$ \quad Mark Yatskar$^{\dagger}$ \quad Chris Callison-Burch$^{\dagger}$ \quad  Andrew Head$^{\dagger}$ \\
Rose Hendrix$^{\dagger}$ \quad Favyen Bastani$^{\dagger}$ \quad Eli VanderBilt$^{\dagger}$ \quad Nathan Lambert$^{\dagger}$ \quad Yvonne Chou$^{\dagger}$\\
Arnavi Chheda$^{\dagger}$ \quad Jenna Sparks$^{\dagger}$ \quad Sam Skjonsberg$^{\dagger}$ \quad  Michael Schmitz$^{\dagger}$ \quad  Aaron Sarnat$^{\dagger}$ \\
Byron Bischoff$^{\dagger}$ \;  Pete Walsh$^{\dagger}$ \; Chris Newell$^{\dagger}$ \; Piper Wolters$^{\dagger}$ \;  Tanmay Gupta$^{\dagger}$ \; Kuo-Hao Zeng$^{\dagger}$\\
Jon Borchardt$^{\dagger}$ \quad  Dirk Groeneveld$^{\dagger}$ \quad  Crystal Nam$^{\dagger}$ \quad Sophie Lebrecht$^{\dagger}$ \quad Caitlin Wittlif$^{\dagger}$ \\ 
Carissa Schoenick$^{\dagger}$ \quad  Oscar Michel$^{\dagger}$ \quad  Ranjay Krishna$^{\dagger\psi}$ \quad  Luca Weihs$^{\dagger}$ \\
Noah A. Smith$^{\dagger\psi}$ \;  Hannaneh Hajishirzi$^{\dagger\psi}$ \;  Ross Girshick$^{\dagger\psi}$ \;  Ali Farhadi$^{\dagger\psi}$ \;  Aniruddha Kembhavi$^{\dagger\psi}$ \\
\vspace*{-2.5mm}\\
$^{\dagger}$Allen Institute for AI \qquad $^{\psi}$University of Washington\vspace{-4.5mm}
}

\let\svthefootnote\thefootnote
\newcommand\freefootnote[1]{%
  \let\thefootnote\relax%
  \footnotetext{#1}%
  \let\thefootnote\svthefootnote%
}

\begin{document}
\maketitle
\freefootnote{$^{*}$Equal contribution}

\begin{abstract}
Today's most advanced vision-language models (VLMs) remain proprietary. The strongest open-weight models rely heavily on synthetic data from proprietary VLMs to achieve good performance, effectively distilling these closed VLMs into open ones. As a result, the community has been missing foundational knowledge about how to build performant VLMs from scratch. We present \textbf{Molmo}, a new family of VLMs that are state-of-the-art in their class of openness. 
Our key contribution is a collection of new datasets called \textbf{PixMo}, including a dataset of highly detailed image captions for pre-training, a free-form image Q\&A dataset for fine-tuning, and an innovative 2D pointing dataset, all collected without the use of external VLMs.
The success of our approach relies on careful modeling choices, a well-tuned training pipeline, and, most critically, the quality of our newly collected datasets. Our best-in-class 72B model not only outperforms others in the class of open weight and data models, but also outperforms larger proprietary models including Claude 3.5 Sonnet, and Gemini 1.5 Pro and Flash, second only to GPT-4o based on both academic benchmarks and on a large human evaluation.
Our model weights, new datasets, and source code are available at \url{https://molmo.allenai.org/blog}.
\end{abstract}
\vspace{-2mm}

\section{Introduction}
\begin{figure*}
    \centering
    \includegraphics[width=\textwidth]{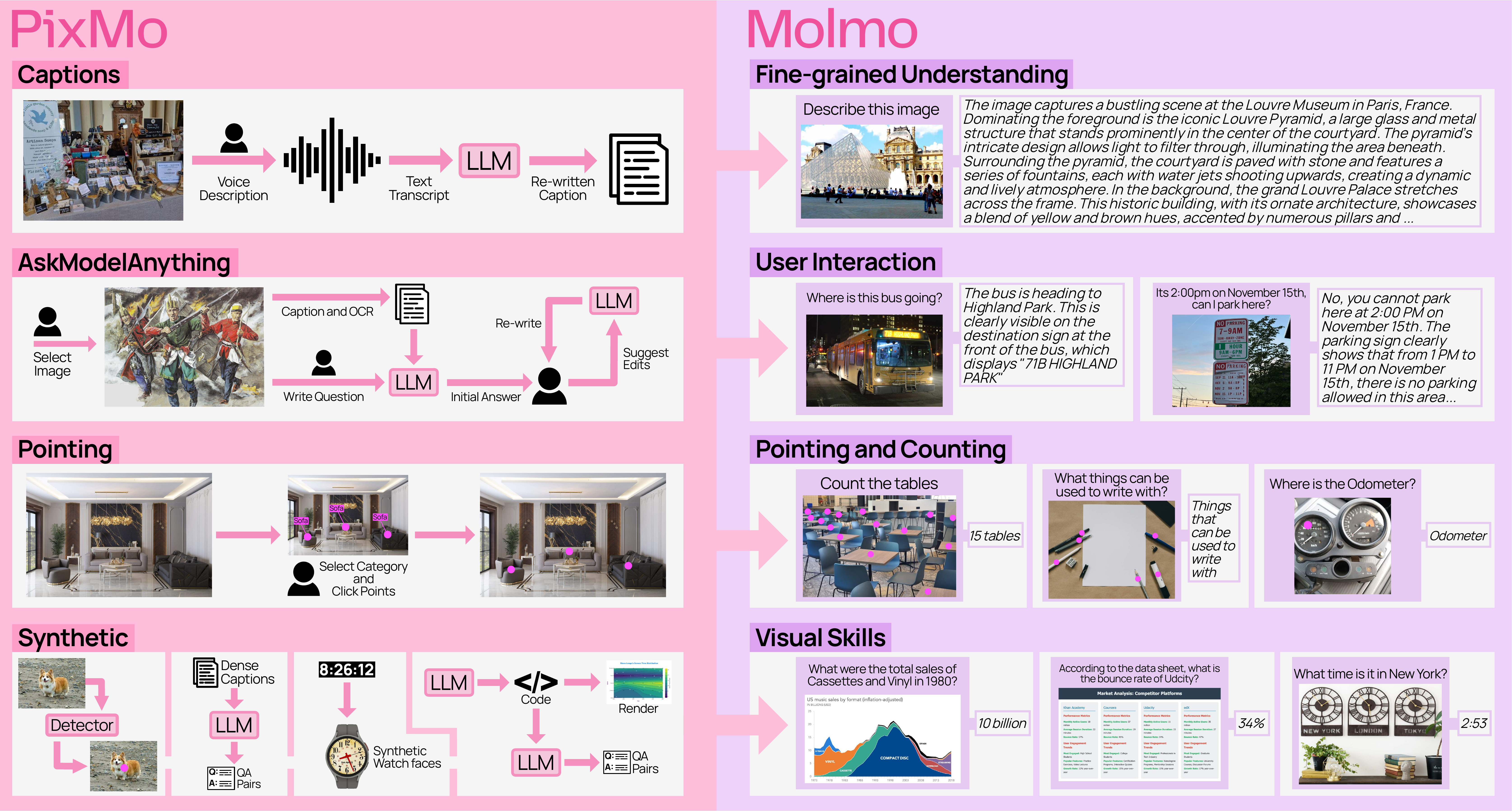}
    \caption{
    Datasets in \textbf{PixMo} (left) and the capabilities they enable in \textbf{Molmo} (right). PixMo consists of three annotated datasets and four synthetic datasets, all constructed without the use of VLMs. The annotated datasets include: dense captions (for pre-training), instruction following (for fine-tuning), and pointing (for fine-tuning, to support grounding and counting). The four synthetic datasets augment these datasets by targeting additional skills (\eg, clock reading, document understanding).
    }
    \label{fig:data}
    \vspace{-5mm}
\end{figure*}
Large multimodal models are used ubiquitously today. Proprietary models—GPT-4o, Gemini-1.5 Pro, Claude 3.5 Sonnet—produce comprehensive image descriptions and accurately answer complex visual questions.
Unfortunately, the most performant of these vision-language models (VLMs) remain proprietary with neither model weights, data, nor code being publicly released.

To foster scientific exploration, numerous research efforts have attempted to reproduce similar capabilities in \emph{open} models.
Early works, exemplified by LLaVA~\cite{liu2023llava}, produced fully open weights and training data but now lag significantly behind the state-of-the-art.
More recent, stronger open-weight models have trended towards less open data: the training data may either be proprietary (\eg,~\cite{beyer2024paligemma}) or, in cases where it is released, there is a heavy reliance on \emph{synthetic} data generated by proprietary systems, \eg, models are trained on datasets like ShareGPT4V~\cite{chen2023sharegpt4v} which uses GPT-4V~\cite{gpt4} to generate a large set of detailed image captions.
The resulting VLMs, therefore, are effectively \emph{distillations} of proprietary VLMs. As it stands, the scientific community is missing foundational knowledge about how to build performant VLMs \emph{from scratch} (more discussion about this and related work are in the Appendix).

In this work, we present the \textbf{Molmo} (\textbf{M}ultimodal \textbf{O}pen \textbf{L}anguage \textbf{Mo}del) family of state-of-the-art open VLMs with released model weights \emph{and} released vision-language training data without any reliance on synthetic data from other VLMs, including proprietary ones.
The success of our approach relies on careful model design choices, a well-tuned training pipeline, and most critically, the quality of our new datasets, collectively named \textbf{PixMo} (\textbf{Pix}els for \textbf{Mo}lmo), which are fully open.

High-quality multimodal data, both for pre-training and fine-tuning, is a key missing piece for training open VLMs that are competitive with closed ones.
The academic community has struggled to collect such datasets due to high costs and the difficulty of obtaining high-quality annotations from crowd sourcing platforms. To build PixMo, we introduce several key data collection innovations that allow us to quickly collect high-quality data from untrained annotators, see Figure~\ref{fig:data}.

PixMo includes a dataset of 712k images with very long (200+ word) detailed captions. Collecting this data was difficult because directly asking annotators to write such captions produces poor results: they tend to focus on a few salient visual elements~\cite{chen2015microsoft}, typing long paragraphs is time-consuming, and annotators can potentially copy-and-paste responses from proprietary VLMs, circumventing our goal of avoiding distillation. Instead, we ask annotators to describe images in \emph{speech} for 60 to 90 seconds.
Empirically, we found that with this modality switching ``trick'' annotators provide far more detailed descriptions in less time, and for each description we collect an audio receipt (\ie, the annotator's recording) proving that a VLM was not used.

PixMo also includes an array of fine-tuning datasets. To collect instruction-following data, we have users interactively edit responses with a language-only LLM to obtain high-quality and accurate free-form responses. We gather 162k annotations on 73k images in this way. 
We also collect a unique new data source that grounds language in images with 2D points.
Using points enables us to collect grounding data much faster than would be possible using bounding boxes or segmentation masks since it is much easier to annotate, and we take advantage of this by collecting over 2.3 million grounding annotations for a diverse range of objects, expressions, and scenes.
This novel pointing data enables our models to answer some questions more naturally by pointing to the pixels that support the answer, improves counting accuracy (the model \emph{counts by pointing}), and we believe it will open up an important future direction in which VLMs enable agents (\eg, robots, web agents) to \emph{act} by pointing in their environments, \eg, to a navigation waypoint, to an object to pick up, or to a user interface button to press.
Finally, we introduce several novel synthetic datasets (meaning with no or minimal human annotations, but still \emph{not} using a VLM) with data targeting particular skills (clock reading, chart understanding, table understanding, \etc) that complements existing open-source datasets.

We train models on these datasets following a mostly standard design using a pre-trained LLM and vision encoder, but with some new improvements including a simplified two-stage training pipeline, a novel overlapping multi-crop strategy, an efficient method of training on images with multiple annotations, and some new insights in how to set up the optimizer and vision/language connector.
We evaluate the Molmo family of models on 11 academic benchmarks and with a human evaluation that allows us to rank models by user preference. Our most efficient model, \mbox{MolmoE-1B}, based on the \mbox{OLMoE-1B-7B}~\cite{muennighoff2024olmoeopenmixtureofexpertslanguage} mixture-of-experts LLM, nearly matches the performance of GPT-4V on both our academic benchmarks and user preference. Molmo-7B-O and Molmo-7B-D, based on \mbox{OLMo-7B}~\cite{OLMo} and Qwen2 7B~\cite{qwen2}, respectively, perform comfortably between GPT-4V and GPT-4o~\cite{hurst2024gpt} on both academic benchmarks and user preference. Our best-in-class Molmo-72B model, based on Qwen2 72B, achieves the highest academic benchmark score and ranks second by human preference, just behind GPT-4o. Our best model outperforms many state-of-the-art proprietary systems, including Gemini 1.5 Pro and Flash~\cite{team2024gemini}, and Claude 3.5 Sonnet~\cite{anthropic2024claude}. We will also release a 100\% fully open Molmo model, based on a MetaCLIP~\cite{xudemystifying} vision encoder and OLMo LLM, for which every bit of training data is publicly available. In addition, we perform an expansive set of ablations to better inform the scientific community of how various model and data design choices affect VLMs.


\begin{figure}[t]\centering
\includegraphics[width=0.98\linewidth]{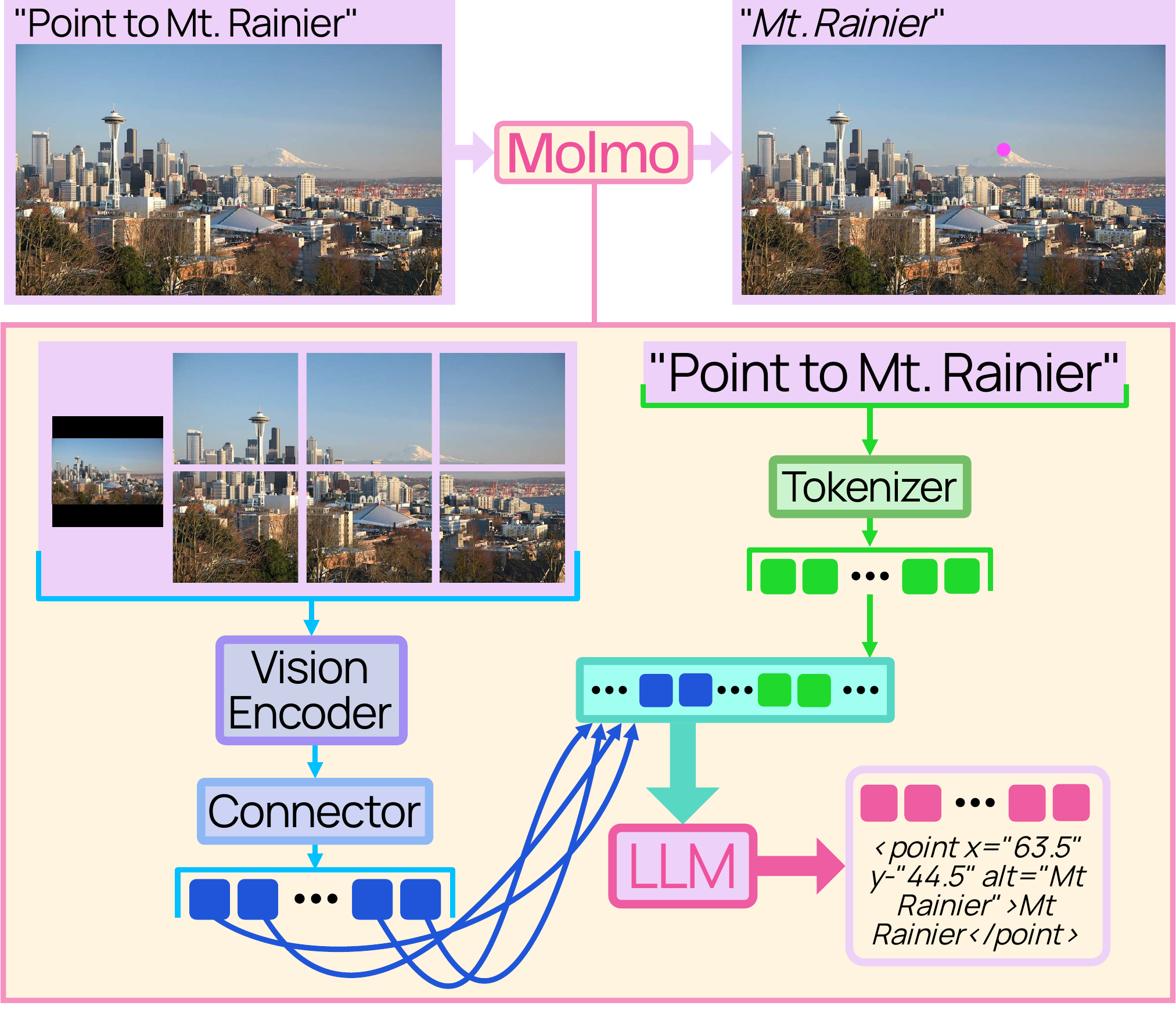}
\caption{\textbf{Molmo} follows the simple and standard design of connecting a vision encoder and a language model.
}\vspace{-4mm}
\label{fig:arch}
\end{figure}

\section{Architecture}
\label{sec:arch}

Our model architecture (Figure~\ref{fig:arch}) follows a standard design, combining pre-trained language and vision models (\eg,~\cite{liu2023llava}). It has four components: (1) a pre-processor that converts the input image into multiscale, multi-crop images, (2) a ViT image encoder~\cite{dosovitskiy2021vit} that computes per-patch features for each image independently, (3) a connector that pools and projects patch features into the LLM's embedding space, and (4) a decoder-only LLM~\cite{vaswani2017attention,radford2018improving}. 

From this template, we build a family of models by selecting a vision encoder and LLM, keeping the training data and recipe consistent across choices (except for learning rates). We primarily use OpenAI's ViT-L/14 336px CLIP model~\cite{clip} due to strong performance in initial experiments, but similar results are achievable with SigLIP~\cite{zhai2023sigmoid} and the \emph{fully} open MetaCLIP~\cite{xudemystifying} (see Section~\ref{sec:ablations}).
For the LLM, we experiment across scales and openness levels: fully open OLMo-7B-1024-preview, fully open OLMoE-1B-7B (our most efficient model), open-weight Qwen2 7B~\cite{qwen2}, and open-weight Qwen2 72B (our best-performing model).

\begin{figure}
    \centering
    \includegraphics[width=0.45\textwidth]{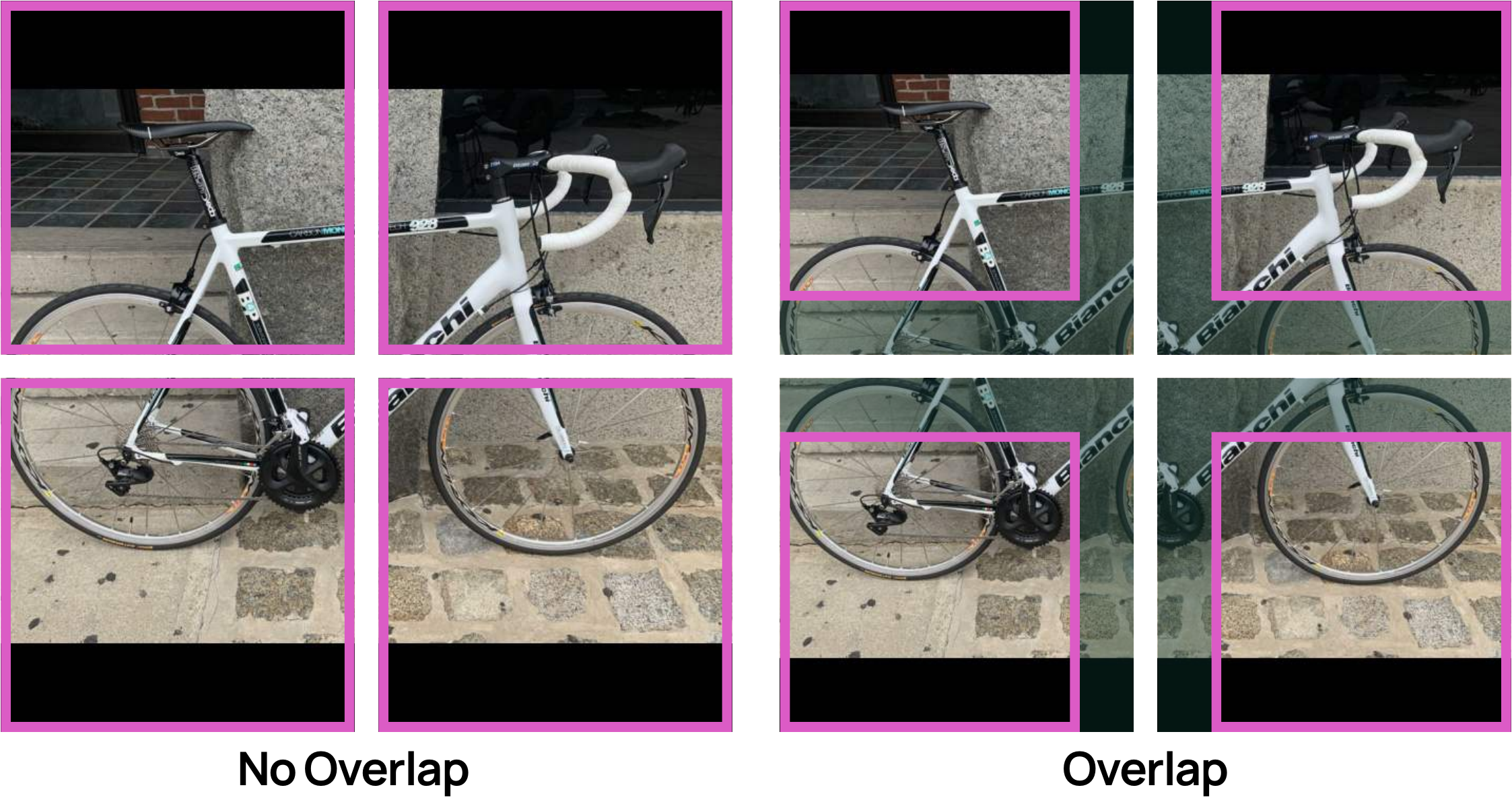}
    \caption{An image cropped without (left) and with (right) overlap. Highlighted regions show areas used by the LLM. Overlapping crops ensure that central patches are encoded with neighboring context; for example, the patches containing the bike's brand name are always part of a crop where the entire name is visible.
    }
    \label{fig:overlapping-crops}
    \vspace{-4mm}
\end{figure}

\paragraph{Cropping.}
Most ViTs only support square images at a fixed resolution that is generally too low for fine-grained tasks such as OCR or detailed captioning.
To address this issue, we follow recent works~\cite{ye2023ureader,dong2024internlm,chen2024far,liu2024improved,McKinzie2024MM1MA} by dividing the image into multiple square crops that tile the image. Additionally, the full image, resized to the ViT's resolution, provides a low-resolution overview. Each crop is processed independently by the ViT. See the Appendix for details.

One limitation of cropping is that border patches lack context from adjacent patches (see Figure~\ref{fig:overlapping-crops}). 
To mitigate this, we allow crops to overlap so each patch has context from at least some neighboring patches.
Patch features from the overlap are \emph{not} passed to the connector or LLM so that the passed patch features exactly tile the high-resolution image. Overlapping slightly reduces the tiled image resolution, but this can be offset by using more crops. Overlapping significantly improves results, as shown in Section~\ref{sec:ablations}.

\paragraph{Vision-language connector.}
Once crops are encoded by the vision encoder, we build patch features by concatenating features from the third-to-last and tenth-from-last ViT layers, which improves performance slightly over using a single layer.
Each 2$\x$2 patch window is then pooled into a single vector using a multi-headed attention layer, where the mean of the patches serves as the query.
This attention pooling outperforms simple feature concatenation (see Section~\ref{sec:ablations}). Finally, pooled features are mapped to the LLM's embedding space via an MLP.

\paragraph{Arranging vision tokens.}
Pooled patch features (vision tokens) are sequenced left-to-right, top-to-bottom, starting with patches from the low-resolution full image, followed by high-resolution crop patches arranged in row-major order. Special tokens are inserted to mark the start and end of both low- and high-resolution patch sequences, with row-end tokens added between rows to indicate row transitions.

\paragraph{Dropout.}
Residual dropout is applied to the LLM, but not the image encoder and vision-language connector. During pre-training on dense captions, dropout is applied only to text tokens to encourage reliance on the encoded image rather than language priors. This is not used in fine-tuning, as shorter target responses result in too little dropout. Text-only dropout during pre-training enhances captioning and downstream performance, as shown in  Section~\ref{sec:ablations}.

\paragraph{Multi-annotated images.}
Our multimodal data often includes multiple annotations per image (\eg, VQA v2.0 has multiple question-answer pairs). To train efficiently, we arrange \emph{all} of text annotation tokens for an image in one long sequence, masking attention so tokens for each annotation attend to the image tokens, each other, but not to tokens from different annotations. This setup is equivalent to training on individual image-text pairs but avoids redundant image encoding, reducing the number of processed images by two-thirds and shortening training time by over half, with only a 25\% increase in sequence length for our data mix.

\section{Data}
\label{sec:data}
\textbf{\mbox{PixMo}} contains seven datasets, three with human annotations and four created with synthetic data generation pipelines (see Figure~\ref{fig:data}). Below, we describe these datasets and data collection methods; additional details and examples are in the Appendix.

\paragraph{PixMo-Cap.} 
We collected PixMo-Cap as a source of high-quality pre-training data, featuring a diverse set of images paired with highly detailed dense captions. We began by sourcing web images across \app70 diverse topics (\eg, street signs, memes, food, drawings, websites, blurry photos, \etc). For each image, three annotators initially provided detailed descriptions by speaking for at least 60 seconds. In later stages, we used one annotator per image with a 90-second minimum, which improved efficiency without sacrificing quality.
We prompted the annotators with seven questions to answer, detailed in the Appendix.

The annotators' audio was transcribed using a standard speech-to-text system, yielding raw transcripts. A final high-quality image caption was then created by prompting a language-only LLM to summarize multiple raw transcripts per image or, for single transcripts, to enhance its quality (\eg, removing spoken artifacts, normalizing style). In total, we collected 712k distinct images with 1.3M transcripts and captions. Our captions average 196 words, compared to 11 words in COCO captions~\cite{chen2015microsoft} and 37 words in localized narratives~\cite{PontTuset_eccv2020}, highlighting their greater detail.

\paragraph{PixMo-AskModelAnything.} We collected this data to enable the model to answer diverse questions it might encounter in real-world use. To create image-question-answer triplets, annotators worked with a language-only LLM. An annotator selected an image from a large pool and wrote a question about it. 
Then, we ran a standard non-VLM OCR model and a PixMo-Cap-trained model on the image. The language-only LLM answered the question from the OCR data and dense caption. The annotator could accept or reject the answer and if rejected, they specified the issue and requested a revision until the answer was satisfactory.
We collected 162k question-answer pairs in 73k images.

\paragraph{PixMo-Points.} We collected pointing data to achieve three goals: (1) enable the model to point to items described by text, (2) enable the model to count by pointing, and (3) use pointing as a form of visual explanation when answering questions. For the first two goals, annotators were asked to point at something in an image, describe it, and then point to each instance of it in the image, ensuring exhaustive coverage. We also collected ``not present'' data so models can learn to handle cases where an item is \emph{not} in the image. Pointing data also naturally supports answering counting questions with a chain-of-thought formed by the sequence of points. This resulted in 2.3M question-points pairs from 223k images. 
To enable points as explanations, we adapted the PixMo-AskModelAnything pipeline to let annotators pass the LLM a list of text-annotated points, prompting the LLM to use them in its answer when relevant. We collected  79k point-explanation annotations on 14k images. 

\paragraph{PixMo-CapQA.} We generated 214k question-answer pairs, covering diverse topics and styles, from 165k images by prompting a language-only LLM to ask and answer questions given only the ground-truth caption for an image.

\paragraph{PixMo-Docs.} We used an extensive and carefully tuned prompting framework to prompt an LLM to generate code for 255k text and figure-heavy images, including charts, documents, tables, and diagrams. We then prompted the LLM to generate 2.3M question-answer pairs based on privileged access to the code (the images were not used). 

\paragraph{PixMo-Clocks.} We rendered synthetic clocks matched with a time-telling question-answer pair. The images use \app50 different watch bodies and \app160k realistic diverse watch faces set to random times. We collected 826k examples.

\paragraph{PixMo-Count.} We used a standard non-VLM object detector~\cite{zhou2022detecting} on web images to create image and counting QA pairs. For each image, we selected the class with the most detections after strict confidence thresholding. Following CountBenchQA~\cite{beyer2024paligemma}, we manually verified 120 samples per count from 2 to 10, creating validation and test sets of 540 images each. These diverse images form a more challenging counting QA set than CountBenchQA, which has reported limitations~\cite{beyer2024paligemma}. The remaining samples with counts between 0 and 10 form a training set of 36k images, each annotated with points (object centers) and a QA pair.

\section{Training}
\paragraph{Pre-training.}
We pre-train all model parameters on PixMo-Cap to generate either the caption or one of the audio transcripts for a given image.
A prompt specifies which style to generate and, 90\% of the time, includes a length hint to guide the model's output length. This hint improves caption and pre-training quality, as shown in Section~\ref{sec:ablations}.

Previous work has often included a separate training stage to tune only the vision-language connector~\cite{li2024llava,tong2024cambrian,dai2024nvlm,liu2023llava,beyer2024paligemma}. 
We find this step unnecessary when pre-training on PixMo-Cap (see Section~\ref{sec:ablations}), also explored in~\cite{karamcheti2024prismatic}.
Instead, we apply a higher learning rate with a shorter warmup for the connector parameters, allowing them to adjust more quickly at the start of training.
Skipping this stage reduces training time and complexity, and eliminates the need for the noisy web-scale data typically used in this phase.

We train for four epochs using AdamW~\cite{kingma2015adam,loshchilov2019decoupled} with a cosine learning rate decaying to 10\% of its peak. Learning rates are set to 2e-4 (connector), 6e-6 (ViT), and 2e-5 (LM), with a 200-step warmup for the connector and 2000 steps for the ViT and LM. Gradient clipping is applied separately to the LM, image encoder, and connector parameters. Full hyper-parameters are provided in the Appendix.

\begin{figure}
    \centering
    \includegraphics[width=0.88\linewidth]{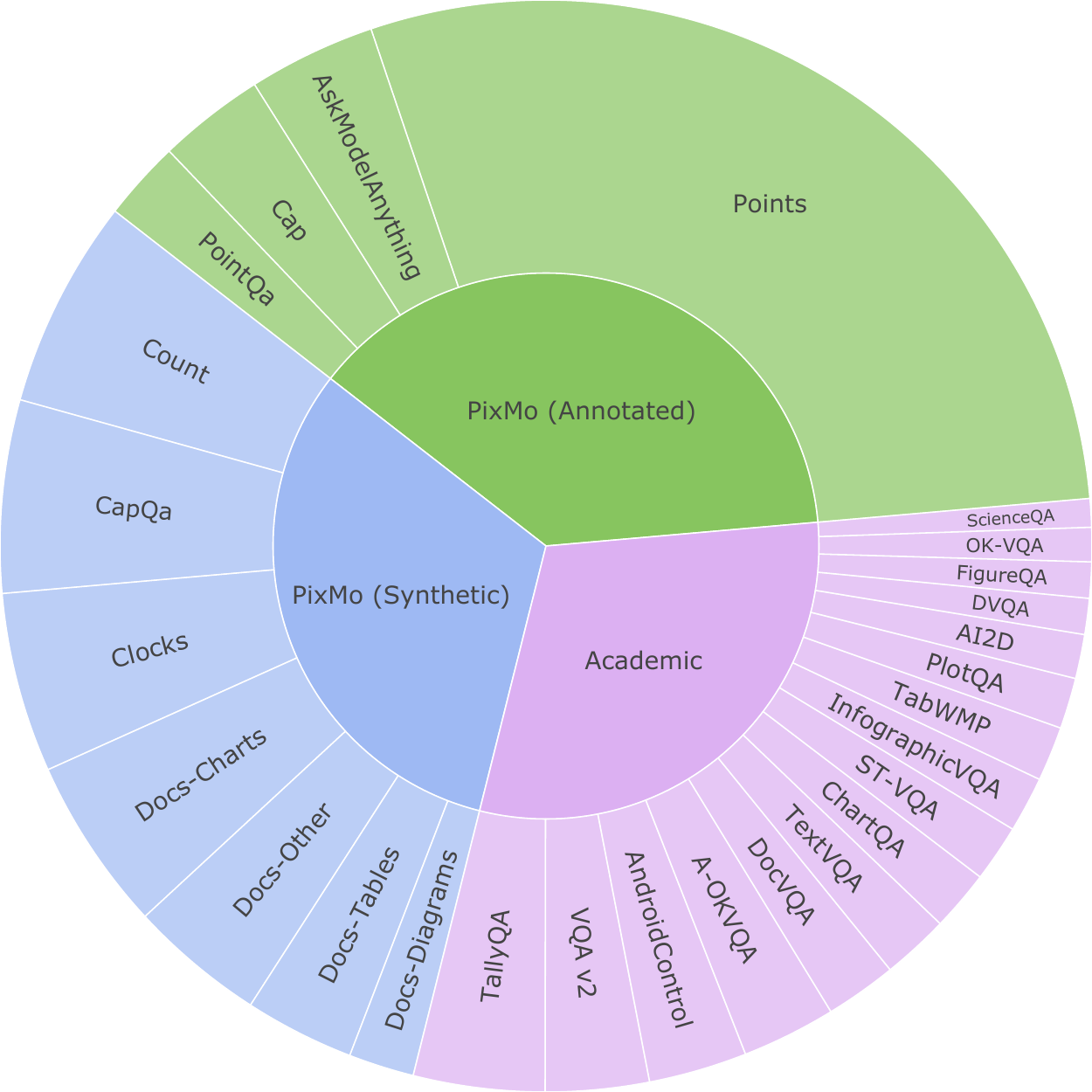}
    \caption{Datasets used for fine-tuning, shown in proportion to their sampling rates. Green denotes human-annotated data we collected, blue denotes synthetic data we generated, and purple represents pre-existing academic datasets. PixMo-Docs has been subdivided into charts, tables, diagrams, and other.}
    \label{fig:mixture}\vspace{-4mm}
\end{figure}

\paragraph{Fine-tuning.}
We fine-tune the model on a mix of PixMo datasets and open-source training datasets, including: VQA v2.0 (COCO 2014 subset)~\cite{goyal2017making}, TextVQA~\cite{singh2019towards}, OK-VQA~\cite{okvqa}, ChartQA (re-weighted to balance human and augmented examples)~\cite{masry-etal-2022-chartqa}, DocVQA~\cite{mathew2021docvqa}, InfographicVQA~\cite{mathew2022infographicvqa}, AI2D (transparent and opaque label boxes)~\cite{kembhavi2016diagram}, A-OKVQA~\cite{schwenk2022okvqa}, AndroidControl~\cite{li2024effects}, ScienceQA~\cite{lu2022learn}, TabMWP~\cite{lu2023dynamic}, ST-VQA~\cite{biten2019scene}, TallyQA~\cite{acharya2019tallyqa}, DVQA~\cite{kafle2018dvqa}, FigureQA~\cite{kahou2017figureqa}, and PlotQA~\cite{methani2020plotqa}.

We sample datasets at rates proportional to the square root of their size, with manual down weighting of some very large synthetic datasets (PlotQA, FigureQA, DVQA, and PixMo-Clocks). We observe that pointing tasks learn more slowly than QA tasks, so we significantly up-weight the pointing data. Final mixture rates are shown in Figure~\ref{fig:mixture}, with full details in the Appendix.

The academic datasets in this mixture teach specific skills and help the model perform well on corresponding benchmark test sets.
However, these datasets often have answer styles that are not ideal for user interactions, as answers are usually very short and may reflect unique stylistic quirks from data collection (\eg, DocQA requires verbatim text from documents, while ChartQA specifies digits without commas).
To prevent these styles from affecting user-facing responses, we prompt the model with a task-specific style tag (\eg, prefixing VQA v2.0 questions with ``vqa2:''). The model learns to use these styles only when requested.

We use style tags for all datasets except PixMo-AskModelAnything, -CapQA, -Points, -Count and -Cap. For PixMo-Cap, we create \app30 prompts for caption generation. For pointing data, we create \app100 question templates that ask for the location or count of the target expression. The model then returns a list of points and, for counting questions, the total count. These prompts and templates are randomly sampled during training.
We still use a style tag for pointing-as-an-explanation data since we find performance in this mode can be less reliable, so it should only be used when users request it.

For pointing, the model outputs points as plain-text coordinates normalized between 0 and 100.
When pointing to multiple items, points are ordered top-down, left-to-right, with each point numbered (see Figure~\ref{fig:arch} and details in the Appendix). 
Pointing enables a unique chain-of-thought approach to counting where the model counts by sequentially pointing to each occurrence of the target object, improving performance (see Section~\ref{sec:ablations}).

\section{Evaluation}
\label{sec:evaluation}
\newcommand{\mycell}[2]{\rotatebox{90}{\parbox{1.7cm}{\setlength{\baselineskip}{0.5em}\textbf{\scriptsize{#1}} \\ \footnotesize{\textcolor{gray}{#2}}}}}
\begin{table*}[!ht]
    \renewcommand{\arraystretch}{0.98}
    \centering
    \small
    \resizebox{\textwidth}{!}{%
    \begin{tabular}{@{}lccccccccccc>{\columncolor{tablegray}}crr@{}}
        \textbf{model} & 
        \mycell{AI2D}{test~\cite{kembhavi2016diagram}}
        & \mycell{ChartQA}{test~\cite{masry-etal-2022-chartqa}}
        & \mycell{VQA v2.0}{testdev~\cite{goyal2017making}}
        & \mycell{DocVQA}{test~\cite{mathew2021docvqa}}
        & \mycell{InfoQA}{test~\cite{mathew2022infographicvqa}}
        & \mycell{TextVQA}{val~\cite{singh2019towards}}
        & \mycell{RealWorldQA}{\cite{realworldqa}}
        & \mycell{MMMU}{val~\cite{yue2024mmmu}}
        & \mycell{MathVista}{testmini~\cite{lu2024mathvista}}
        & \mycell{CountBenchQA}{\cite{beyer2024paligemma}}
        & \mycell{PixMo-Count}{test}
        & \mycell{Average}{}
        & \mycell{Elo score}{}
        & \mycell{Elo rank}{} \\
        \midrule
        
        \multicolumn{12}{@{}l}{\textbf{\textit{API call only}}} & \cellcolor{tablegray}\\
        GPT-4V~\cite{gpt4} & 89.4 & 78.1 & 77.2 & 87.2 & 75.1 & 78.0 & 61.4 & 63.1 & 58.1 & 69.9 & 45.0 & 71.1 & 1041 & 10 \\
        GPT-4o-0513~\cite{hurst2024gpt} & 94.2 & 85.7 & 78.7 & 92.8 & 79.2 & 77.4 & 75.4 & \textbf{69.1} & 63.8 & 87.9 & 59.6 & 78.5 & \textbf{1079} & \textbf{1}\\

        Gemini 1.5 Flash~\cite{team2024gemini} & 91.7 & 85.4 & 80.1 & 89.9 & 75.3 & 78.7 & 67.5 & 56.1 & 58.4 & 81.6 & 61.1 & 75.1 & 1054 & 7 \\ 
        Gemini 1.5 Pro~\cite{team2024gemini} & 94.4 & 87.2 & 80.2 & 93.1 & 81.0 & 78.7 & 70.4 & 62.2 & 63.9 & 85.8 & 64.3 & 78.3 & 1074 & 3\\ 

        Claude-3 Haiku~\cite{anthropic2024claude} & 86.7 & 81.7 & 68.4 & 88.8 & 56.1 & 67.3 & 45.5 & 50.2 & 46.4 & 83.0 & 43.9 & 65.3 & 999 & 18 \\ 
        Claude-3 Opus~\cite{anthropic2024claude} & 88.1 & 80.8 & 66.3 & 89.3 & 55.6 & 67.5 & 49.8 & 59.4 & 50.5 & 83.6 & 43.3 & 66.7 & 971 & 21 \\
        Claude-3.5 Sonnet ~\cite{anthropic2024claude} & 94.7 & \textbf{90.8} & 70.7 & 95.2 & 74.3 & 74.1 & 60.1 & 68.3 & \textbf{67.7} & 89.7 & 58.3 & 76.7 & 1069 & 4\\ 
        \midrule

        \multicolumn{12}{@{}l}{\textbf{\textit{Open weights only}}} & \cellcolor{tablegray}\\
        PaliGemma-mix-3B~\cite{beyer2024paligemma} & 72.3 & 33.7 & 76.3 & 31.3 & 21.4 & 56.0 & 55.2 & 34.9 & 28.7 & 80.6 & 60.0 & 50.0 & 937 & 27 \\
        Phi3.5-Vision-4B~\cite{abdin2024phi} & 78.1 & 81.8 & 75.7 & 69.3 & 36.6 & 72.0 & 53.6 & 43.0 & 43.9 & 64.6 & 38.3 & 59.7 & 982 & 19 \\
        Qwen2-VL-7B~\cite{wang2024qwen2} & 83.0 & 83.0 & 82.9 & 94.5 & 76.5 & 84.3 & 70.1 & 54.1 & 58.2 & 76.5 & 48.0 & 73.7 & 1025 & 14 \\
        Qwen2-VL-72B~\cite{wang2024qwen2} & 88.1 & 88.3 & 81.9 & \textbf{96.5} & 84.5 & \textbf{85.5} & \textbf{77.8} & 64.5 & 70.5 & 80.4 & 55.7 & 79.4 & 1037 & 12 \\
        InternVL2-8B~\cite{internvl2} & 83.8 & 83.3 & 76.7 & 91.6 & 74.8 & 77.4 & 64.2 & 51.2 & 58.3 & 57.8 & 43.9 & 69.4 & 953 & 23 \\
        InternVL2-Llama-3-76B~\cite{internvl2} & 87.6 & 88.4 & 85.6 & 94.1 & \textbf{82.0} & 84.4 & 72.7 & 58.2 & 65.5 & 74.7 & 54.6 & 77.1 & 1018 & 16 \\
        Pixtral-12B~\cite{agrawal2024pixtral} & 79.0 & 81.8 & 80.2 & 90.7 & 50.8 & 75.7 & 65.4 & 52.5 & 58.0 & 78.8 & 51.7 & 69.5 & 1016 & 17 \\ 
        Llama-3.2V-11B-Instruct~\cite{llama3} & 91.1 & 83.4 & 75.2 & 88.4 & 63.6 & 79.7 & 64.1 & 50.7 & 51.5 & 73.1 & 47.4 & 69.8 & 1040 & 11 \\
        Llama-3.2V-90B-Instruct~\cite{llama3} & 92.3 & 85.5 & 78.1 & 90.1 & 67.2 & 82.3 & 69.8 & 60.3 & 57.3 & 78.5 & 58.5 & 74.5 & 1063 & 5 \\
        
        \midrule

        \multicolumn{12}{@{}l}{\textbf{\textit{Open weights + data \;($\dagger$ distilled)}}} & \cellcolor{tablegray}\\
        LLaVA-1.5-7B~\cite{liu2023llava} & 55.5 & 17.8 & 78.5 & 28.1 & 25.8 & 58.2 & 54.8 & 35.7 & 25.6 & 40.1 & 27.6 & 40.7 & 951 & 26 \\
        LLaVA-1.5-13B~\cite{liu2023llava} & 61.1 & 18.2 & 80.0 & 30.3 & 29.4 & 61.3 & 55.3 & 37.0 & 27.7 & 47.1 & 35.2 & 43.9 & 960 & 22 \\
        xGen-MM-interleave-4B$\dagger$~\cite{xue2024xgen} & 74.2 & 60.0 & 81.5 & 61.4 & 31.5 & 71.0 & 61.2 & 41.1 & 40.5 & 81.9 & 50.2 & 59.5 & 979 & 20 \\
        Cambrian-1-8B$\dagger$~\cite{tong2024cambrian} & 73.0 & 73.3 & 81.2 & 77.8 & 41.6 & 71.7 & 64.2 & 42.7 & 49.0 & 76.4 & 46.6 & 63.4 & 952 & 25 \\
        Cambrian-1-34B$\dagger$~\cite{tong2024cambrian} & 79.7 & 75.6 & 83.8 & 75.5 & 46.0 & 76.7 & 67.8 & 49.7 & 53.2 & 75.6 & 50.7 & 66.8 & 953 & 24 \\
        LLaVA OneVision-7B$\dagger$~\cite{li2024llava} &81.4 & 80.0 & 84.0 & 87.5 & 68.8 & 78.3 & 66.3 & 48.8 & 63.2 & 78.8 & 54.4 & 72.0 & 1024 & 15 \\
        LLaVA OneVision-72B$\dagger$~\cite{li2024llava} & 85.6 & 83.7 & 85.2 & 91.3 & 74.9 & 80.5 & 71.9 & 56.8 & 67.5 & 84.3 & 60.7 & 76.6 & 1051 & 8 \\
        \midrule

        \multicolumn{12}{@{}l}{\textbf{\textit{The Molmo family: Open weights, Open data, Open training code, Open evaluations}}} & \cellcolor{tablegray}\\
        \textcolor{molmocolor}{MolmoE-1B} & 86.4 & 78.0 & 83.9 & 77.7 & 53.9 & 78.8 & 60.4 & 34.9 & 34.0 & 87.2 & 79.6 & 68.6 & 1032 & 13 \\
        \textcolor{molmocolor}{Molmo-7B-O} & 90.7 & 80.4 & 85.3 & 90.8 & 70.0 & 80.4 & 67.5 & 39.3 & 44.5 & 89.0 & 83.3 & 74.6 & 1051 & 9 \\
        \textcolor{molmocolor}{Molmo-7B-D} & 93.2 & 84.1 & 85.6 & 92.2 & 72.6 & 81.7 & 70.7 & 45.3 & 51.6 & 88.5 & 84.8 & 77.3 & 1056 & 6 \\
        \textcolor{molmocolor}{Molmo-72B} & \textbf{96.3} & 87.3 & \textbf{86.5} & 93.5 & 81.9 & 83.1 & 75.2 & 54.1 & 58.6 & \textbf{91.2} & \textbf{85.2} & \textbf{81.2} & 1077 & 2\\
    \end{tabular}
    }%
    \caption{We present academic benchmark results for 10 common datasets, plus a new counting benchmark, PixMo-Count, which features more challenging natural images than CountBenchQA. We categorize models into four groups: (top) proprietary models accessible only via API calls, (upper middle) models with released weights but closed data, (lower middle) models with released weights and training data (noting some of these use distillation ($\dagger$) from proprietary VLMs via synthetic data), and (bottom) the Molmo family of models.}\vspace{-4mm}
    \label{tab:benchmark_results}
\end{table*}

\begin{table*}[th!]
\vspace{-.2em}
\centering
\subfloat[
\textbf{Vision encoder}. Encoders that were trained on noisy web-scale image-text pairs perform similarly (rows 1-3). Surprisingly, DINOv2, which is trained on images only (no text, no label supervision), is competitive on these tasks. MetaCLIP and DINOv2 are \emph{fully} open.
\label{tab:vision_encoder}
]{
\begin{minipage}{0.29\linewidth}{\begin{center}
\tablestyle{2pt}{1.02}
\begin{tabular}{y{74}x{24}x{24}}
ViT-L/14 & cap $F_{1}$ & 11-avg \\
\shline
\baseline{OpenAI CLIP 336px} & 54.1 & 76.9 \\
MetaCLIP 336px & 54.1 & \textbf{77.2} \\
SigLIP-So400m 384px & \textbf{54.4} & 77.1 \\
DINOv2 336px & 53.2 & 75.6 \\
\multicolumn{3}{c}{~}\\
\multicolumn{3}{c}{~}\\
\end{tabular}
\end{center}}\end{minipage}
}
\hspace{2em}
\subfloat[
\textbf{Image resolution}. Using more crops at training and testing time generally improves performance. However, captioning and counting can perform poorly when $\#$ of crops are unequal, so for these tasks we always set the number of test crops equal to the training value.
\label{tab:resolution}
]{
\begin{minipage}{0.29\linewidth}{\begin{center}
\tablestyle{2pt}{1.02}
\begin{tabular}{y{60}x{24}x{24}}
\# crops train, test & cap $F_{1}$ & 11-avg \\
\shline
\phantom{1}4, 4 & 52.0 & 71.0 \\
\phantom{1}4, 12\phantom{*} & 52.0 & 74.1 \\
\phantom{1}4, 36\phantom{*} & 52.0 & 74.2 \\
12, 12 & 54.1 & 74.9 \\
\baseline{12, 36} & \textbf{54.1} & 76.9 \\
36, 36 & 54.0 & \textbf{77.2} \\
\end{tabular}
\end{center}}\end{minipage}
}
\hspace{2em}
\subfloat[
\textbf{Dropout}. Dropout in the LLM improves pre-training and fine-tuning results. In pre-training, applying dropout to captioning text tokens only further improves results. This design may encourage the model to rely more on vision tokens rather than past text tokens.
\label{tab:dropout}
]{
\begin{minipage}{0.29\linewidth}{\begin{center}
\tablestyle{2pt}{1.02}
\begin{tabular}{y{64}x{24}x{24}}
pre-train, fine-tune & cap $F_{1}$ & 11-avg \\
\shline
off, off & 53.1 & 74.6 \\
off, on & 53.1 & 76.6 \\
on, on & 53.7 & \textbf{77.0} \\
\baseline{on (text only), on} & \textbf{54.1} & 76.9 \\
\multicolumn{3}{c}{~}\\
\multicolumn{3}{c}{~}\\
\end{tabular}
\end{center}}\end{minipage}
}
\\
\centering
\vspace{-.5em}
\subfloat[
\textbf{Cropping}. Using the entire image only (single crop) performs poorly. Our novel overlapping crop method (see Figure~\ref{fig:overlapping-crops}), which prevents loss of context, performs the best.
\label{tab:cropping}
]{
\centering
\begin{minipage}{0.29\linewidth}{\begin{center}
\tablestyle{2pt}{1.02}
\begin{tabular}{y{60}x{24}x{24}}
cropping & cap $F_{1}$ & 11-avg \\
\shline
single & 46.7 & 62.8 \\
multi, no overlap & 53.4 & 75.7 \\
\baseline{multi, overlap} & \textbf{54.1} & \textbf{76.9} \\
\end{tabular}
\end{center}}\end{minipage}
}
\hspace{2em}
\subfloat[
\textbf{Length conditioning}. Captioning with length hints is a superior pre-training task compared to captioning alone as evident by the improved captioning and downstream results.
\label{tab:len_cond}
]{
\centering
\begin{minipage}{0.29\linewidth}{\begin{center}
\tablestyle{2pt}{1.02}
\begin{tabular}{y{54}x{24}x{24}}
setting & cap $F_{1}$ & 11-avg \\
\shline
off & 53.0 & 76.2 \\
\baseline{on} & \textbf{54.1} & \textbf{76.9} \\
\multicolumn{3}{c}{~}\\
\end{tabular}
\end{center}}\end{minipage}
}
\hspace{2em}
\subfloat[
\textbf{Pooling}. Pooling 2$\x$2 windows of vision tokens using mean-query attention performs better than simply stacking the four features as input to the vision-language connector MLP.
\label{tab:pooling}
]{
\begin{minipage}{0.29\linewidth}{\begin{center}
\tablestyle{2pt}{1.02}
\begin{tabular}{y{56}x{24}x{24}z{24}}
2$\x$2 pooling & cap $F_{1}$ & 11-avg \\
\shline
stacking & 53.7 & 76.1 \\
\baseline{attention} & \textbf{54.1} & \textbf{76.9} \\
\multicolumn{3}{c}{~}\\
\end{tabular}
\end{center}}\end{minipage}
}
\caption{\textbf{Model ablations.} Default settings are marked in \colorbox{baselinecolor}{gray}. See the Appendix for additional ablations.}
\label{tab:model_ablations} \vspace{-.5em}
\end{table*}

We evaluate on academic benchmarks, noting that comparisons require care, as prompting, alignment with benchmark-specific answer styles, and use of benchmark training data can significantly affect performance.
To complement this, we conduct a human evaluation to rank models based on user preference.

For academic benchmarking, we gather or compute results for all models on 10 common datasets and the PixMo-Count test set, which we include due to its higher difficulty compared to existing counting benchmarks.
We prioritize author-published results but fill in missing results with the best previously reported values from technical reports or sources like the OpenVLM Leaderboard. If data is still missing, we compute it ourselves. Notably, computing results is challenging, as performance can vary significantly (\eg, by 10\%) based on evaluation details. Additionally, critical information such as prompts or data processing steps is often unavailable, making it hard to reproduce results, highlighting the need for evaluation openness.

In our human evaluation, we collect 15k diverse image-text prompt pairs and queried the VLMs for responses. We sample and present the resulting image-text-response triplets for all VLM pairings to a group of \app870 human annotators, who provide pairwise preference rankings. Across all model pairs, we gather over 325k ratings (\app450 per model pair). From this data, we calculate an Elo ranking using the Bradley-Terry model, following the methodology of Chatbot Arena~\cite{chiang2024chatbot}.

For Molmo, we evaluate all academic datasets with 36 crops (up from 12 used in training), except for counting tasks, as pointing capabilities do not generalize well with different numbers of test crops. A small amount of high-res post-training can resolve this issue, see the Appendix.

When possible, we use relevant style prompts\footnote{We use AI2D with transparent boxes; see Appendix for opaque boxes.} (\eg, ``vqa2:''). For evaluation-only datasets, we use the VQA v2.0 (for short answer) or A-OKVQA (for multiple choice) style tags to elicit the often expected short answer style. For human evaluation, we omit style tags and use 12 crops, as some counting questions use pointing. Evaluators are only shown the output text, not the points.

Broadly speaking, the academic benchmark results and human evaluation agree, with the exception of Qwen2-VL~\cite{wang2024qwen2}, which performs strongly on the academic benchmarks and comparatively underperforms in the human evaluation. We highlight a few key results from Table~\ref{tab:benchmark_results}:
\begin{itemize}[leftmargin=5mm,topsep=1mm]
    \item MolmoE-1B, our most efficient model based on the OLMoE-1B-7B mixture-of-experts LLM, nearly matches GPT-4V on academic benchmarks and Elo.
    \item Molmo models based on OLMo-7B-1024-preview and Qwen2 7B LLMs perform between GPT-4V and GPT-4o on academic benchmarks and Elo.
    \item Our best-in-class Qwen2 72B based model achieves the highest academic benchmark score and ranks second in Elo, just behind GPT-4o.
    \item Our best model also outperforms many state-of-the-art proprietary systems, including Gemini 1.5 Pro and Flash and Claude 3.5 Sonnet.
\end{itemize}

Molmo-72B also underwent an independent Elo evaluation via Chatbot Arena, where it outperforms all open models but ranks lower than several proprietary models (\eg., GPT-4o and Claude 3.5 Sonnet).\footnote{\url{https://lmarena.ai/?leaderboard} vision arena, English category, accessed Nov. 13, 2024.} The full results table is in the Appendix. The difference likely stems from the types of questions evaluated. While we cannot perform a full analysis since the questions are not public, we do note our data includes many counting and image-description questions which are particular strengths of Molmo.

Molmo excels at answering questions about natural images, matching or outperforming all models on the zero-shot RealWorldQA benchmark and achieving state-of-the-art results on the highly competitive VQA v2.0. On OCR-centric benchmarks (ChartQA, DocQA, InfoQA, TextVQA), Molmo surpasses other open models and some proprietary ones but trails slightly behind Qwen2-VL. On counting tasks (CountBenchQA and PixMo-Count), Molmo leads all models due to our new pointing data and chain-of-thought point-and-count abilities. However, on reasoning tasks (MMMU, MathVista) Molmo lags, likely because its training mix lacks data focused on advanced reasoning.

We conduct several additional skill-specific evaluations, summarized here with details in the Appendix.
On a clock-reading benchmark~\cite{yang2022s}, Molmo at all scales dramatically outperforms other VLMs including proprietary ones, but trails specialized non-VLM models~\cite{yang2022s}. 
To assess Molmo’s potential for \emph{action}, we tested Molmo-72B on AndroidControl~\cite{li2024effects}, achieving 88.7\% low-level and 69.0\% high-level accuracy, comparable to the reported 83.2\% and 70.8\% in~\cite{li2024effects}.
On NLP benchmarks, Molmo shows a slight performance drop versus its component LLM, which can be offset by additional text-only data.
We also introduce a new pointing benchmark using SAM~\cite{kirillov2023segment}, where Molmo models at all scales demonstrate strong performance.

\begin{table*}[t]
\vspace{-.0em}
\centering
\subfloat[
\textbf{PixMo-Cap scaling}. Increasing the quantity of PixMo-Cap captioning data in both pre-training and fine-tuning improves captioning and downstream tasks.
\label{tab:scaling}
]{
\begin{minipage}{0.29\linewidth}{\begin{center}
\tablestyle{2pt}{1.02}
\begin{tabular}{y{68}x{24}x{24}}
\# PixMo-Cap images & cap $F_{1}$ & 11-avg \\
\shline
\phantom{00}0\phantom{k} (0.0\%) & - & 74.9 \\
\phantom{0}89k (12.5\%) & 49.6 & 75.5 \\
178k (25.0\%) & 51.6 & 76.3 \\
356k (50.0\%) & 52.6 & 76.2 \\
\baseline{712k (100.0\%)} & \textbf{54.1} & \textbf{76.9} \\
\multicolumn{3}{c}{~}\\
\multicolumn{3}{c}{~}\\
\multicolumn{3}{c}{~}\\
\end{tabular}
\end{center}}\end{minipage}
}
\hspace{2em}
\subfloat[
\textbf{Pre-training data}. Our human annotated data performs on par with distilling GPT-4o by using it to caption the same set of images, demonstrating the effectiveness of our data.
\label{tab:stage1}
]{
\begin{minipage}{0.29\linewidth}{\begin{center}
\tablestyle{2pt}{1.02}
\begin{tabular}{y{100}x{24}x{24}}
data & cap $F_{1}$ & 11-avg \\
\shline
stage 0.5 LAION & 53.9 & 76.9 \\
ShareGPT4V+o (158k images) & 36.3 & 74.9 \\
\emph{PixMo-Cap images:} & \multicolumn{2}{c}{~} \\
~~our raw transcripts only & 45.2 & 76.4 \\
~~our cleaned transcripts only & 53.0 & 76.5 \\
~~\baseline{our raw \& cleaned transcripts} & \textbf{54.1} & 76.9 \\
~~captioned by GPT-4o & 52.9 & \textbf{77.5} \\
\multicolumn{3}{c}{~}\\
\end{tabular}
\end{center}}\end{minipage}
}
\hspace{2em}
\subfloat[
\textbf{Supervised fine-tuning data}. The \mbox{PixMo-$\star$} datasets not only give the model new capabilities (\eg, pointing), but also generally improve results on the 11 dataset benchmark.
\label{tab:stage2}
]{
\begin{minipage}{0.29\linewidth}{\begin{center}
\tablestyle{2pt}{1.02}
\begin{tabular}{y{100}x{24}}
data & 11-avg \\
\shline
academic only & 72.5 \\
~~plus PixMo-Docs & 74.0 \\
\baseline{PixMo-$\star$ plus academic} & 76.9 \\
~~remove PixMo-AMA & 76.8 \\
~~remove PixMo-CapQA & \textbf{77.0} \\
~~remove PixMo-Docs & 75.8 \\
~~remove PixMo-Clocks & 76.9 \\
~~remove pointing task & 76.2 \\
\end{tabular}
\end{center}}\end{minipage}
}
\caption{\textbf{Data ablations.} Default settings are marked in \colorbox{baselinecolor}{gray}.}
\label{tab:data_ablations} \vspace{-.5em}
\end{table*}

\begin{table*}[t]
\vspace{-.2em}
\centering
\subfloat[
\textbf{Counting strategy}. Pointing is the key ingredient in Molmo's counting abilities. 
\label{tab:counting_strategy}
]{
\begin{minipage}{0.22\linewidth}{\begin{center}
\tablestyle{2pt}{1.02}
\begin{tabular}{lcc}
strategy & CBQA & PCQA \\
\hline
count & 87.9 & 80.2 \\
\baseline{point then count} & \textbf{89.4} & \textbf{86.3} \\
count then point & 81.5 & 77.6 \\
pointing + regex & 88.4 & 85.4 \\
\end{tabular}
\end{center}}\end{minipage}
}\hfill
\subfloat[
\textbf{Point order}. Training on ordered points (top-down, left-right) is better than unordered points.
\label{tab:counting_random_order}
]{
\begin{minipage}{0.22\linewidth}{\begin{center}
\tablestyle{2pt}{1.02}
\begin{tabular}{lcc}
order & CBQA & PCQA \\
\hline
\baseline{on} & \textbf{89.4} & \textbf{86.3} \\
off & 85.4 & 74.1 \\
\multicolumn{3}{c}{~}\\
\multicolumn{3}{c}{~}\\
\end{tabular}
\end{center}}\end{minipage}
}\hfill
\subfloat[
\textbf{Inference compute}. Simply increasing inference compute with extra tokens does not help.
\label{tab:counting_random_points}
]{
\begin{minipage}{0.22\linewidth}{\begin{center}
\tablestyle{2pt}{1.02}
\begin{tabular}{lcc}
points, length & CBQA & PCQA \\
\hline
\baseline{actual, correct} & \textbf{89.4} & \textbf{86.3} \\
random, correct& 85.9 & 76.3 \\
random, random & 76.3 & 75.7 \\
\multicolumn{3}{c}{~}\\
\end{tabular}
\end{center}}\end{minipage}
}\hfill
\subfloat[
\textbf{Special point tokens}. Encoding point coordinates as plain-text works best.
\label{tab:counting_special_tokens}
]{
\begin{minipage}{0.22\linewidth}{\begin{center}
\tablestyle{2pt}{1.02}
\begin{tabular}{lcc}
tokens & CBQA & PCQA \\
\hline
\baseline{plain-text} & \textbf{89.4} & \textbf{86.3} \\
special & 85.8 & 80.9 \\
\multicolumn{3}{c}{~}\\
\multicolumn{3}{c}{~}\\
\end{tabular}
\end{center}}\end{minipage}
}
\caption{\textbf{Counting ablations.} Defaults are in \colorbox{baselinecolor}{gray}. CBQA is the CountBenchQA test set and PCQA is the PixMo-Count validation set.}\vspace{-4mm}
\end{table*}

\section{Ablations}
\label{sec:ablations}
We performed extensive ablations on model design (Table~\ref{tab:model_ablations}) and training data (Table~\ref{tab:data_ablations}), reporting two metrics: an $F_{1}$ metric (``cap $F_{1}$'') we developed to measure the precision and recall of captions generated by the model after pre-training (details in the Appendix), and the average accuracy on our 11 benchmark suite (``11-avg''), using validation sets when available. 
We report $F_{1}$ because we believe it reflects broad-range image understanding learned during pre-training, and many of our modeling design choices were based on this evaluation since it does not require running the more costly fine-tuning stage. While performing these ablation we observed captioning improvements generally, but not always, correspond to benchmark suite improvements.
Our ablations test modifications to the Molmo-7B-D model configuration with key findings summarized below and further details in Table~\ref{tab:model_ablations} and~\ref{tab:data_ablations} captions and the Appendix.

\paragraph{Model ablations.} We vary several design choices, finding:
\begin{itemize}[leftmargin=5mm,topsep=1mm]
    \item Vision encoders trained on noisy web-scale data (CLIP, SigLIP, MetaCLIP) all work roughly the same, including the \emph{fully} open MetaCLIP (Table~\ref{tab:vision_encoder}). \emph{Every model component and every bit of data in a Molmo model equipped with MetaCLIP and OLMo is open.}
    \item Excluding pointing and captioning, increasing the image resolution by increasing the number of crops generally improves performance, and tuning at a high resolution yields a slight gain (Table~\ref{tab:resolution}).
    \item Our novel text-only dropout used in pre-training improves captioning performance (Table~\ref{tab:dropout}).
    \item Using multiple crops, instead of a single low-resolution input, is extremely important. Our novel overlapping crop design yields significant improvements (Table~\ref{tab:cropping}).
    \item Our novel length-conditioned captioning is a strong pre-training task, improving downstream results (Table~\ref{tab:len_cond}).
    \item Attention pooling yields improvements to both metrics over the baseline feature stacking approach (Table~\ref{tab:pooling}).
\end{itemize}

\paragraph{Data ablations.} We train with various data choices, finding:
\begin{itemize}[leftmargin=5mm,topsep=1mm]
    \item Scaling PixMo-Cap from 0 to 712k images significantly improves captioning and benchmark metrics (Table~\ref{tab:scaling}).
    \item Adding noisy web-scale data does not improve our pre-training, our PixMo-Cap data yields better results than a similar amount of data from ShareGPT4v/o, and our human annotated captioning data works roughly as well as captioning data generated by GPT-4o on the PixMo-Cap images (Table~\ref{tab:stage1}).
    \item The PixMo supervised fine-tuning datasets improve benchmark task performance beyond the academic datasets, mainly by improving document-heavy and counting tasks (Table~\ref{tab:stage2}).
\end{itemize}

\paragraph{Counting.} We ablate several details of counting using models fine-tuned on just PixMo-Points and PixMo-Count data:
\begin{itemize}[leftmargin=5mm,topsep=1mm]
    \item Chain-of-thought point-then-count performs significantly better than generating a count only or generating a count followed by points. Pointing on its own, counted by regular expression, is only slightly worse (Table~\ref{tab:counting_strategy}).
    \item Training with points in a predicable spatial order (top-down, left-to-right) works significantly better than training with unordered points (Table~\ref{tab:counting_random_order}).
    \item Pointing uses extra inference computation that may improve results regardless of the generated text. However, we find that using correct point locations and counts works significantly better than using random locations and/or numbers of points (Table~\ref{tab:counting_random_points}).
    \item Representing point coordinates in plain-text works better than introducing special location tokens (Table~\ref{tab:counting_special_tokens}).
\end{itemize}

\begin{table}[t]
\centering
\tablestyle{2pt}{1.05}
\resizebox{\linewidth}{!}{%
\begin{tabular}{@{}lx{32}c@{}}
model & Elo score & win \% \vs \colorbox{baselinecolor}{default} \\
\shline
\emph{Claude-3.5 Sonnet} & \textbf{1047} & \textbf{65\%} \\
PixMo-Cap w/ GPT-4o captions (Table~\ref{tab:stage1}) & 1018 & 55\% \\
PixMo-$\star$, remove PixMo-CapQA (Table~\ref{tab:stage2}) & 1015 & 50\% \\
\baseline{Molmo-7B-D default} & 1014 & n/a \\
PixMo-$\star$, no academic datasets (Table~\ref{tab:stage2}) & 1013 & 42\% \\
\emph{GPT-4V} & 1010 & 47\% \\
DINOv2 vision encoder (Table~\ref{tab:vision_encoder}) & \phantom{1}999 & 45\% \\
PixMo-$\star$, remove PixMo-AMA (Table~\ref{tab:stage2}) & \phantom{1}995 & 40\% \\
no PixMo-Cap data (Table~\ref{tab:scaling}) & \phantom{1}990 & 35\% \\
academic only (Table~\ref{tab:stage2}) &  \phantom{1}897 & 17\% \\
\end{tabular}
}%
\caption{\textbf{Elo} scores and win rates (excluding ties) for select ablations of Molmo-7B-D and two \emph{API-only} models for context.}
\label{tab:ablation_elo}\vspace{-4mm}
\end{table}

\paragraph{Human evaluation.} A human evaluation of select ablation models in Table~\ref{tab:ablation_elo} shows that PixMo data, especially PixMo-Cap and PixMo-AskModelAnything, is important for generating responses that users like. 
Academic datasets improve human scores, but are extremely poor if used on their own. 
GPT-4o captions on our images also perform well, which we believe is due to recent advances in GPT and the diversity of our image collection (\eg, the ShareGPT datasets significantly underperform our data even at the same scale, see Tables~\ref{tab:stage1} and~\ref{tab:scaling}).
While distilling from proprietary models might be effective, we emphasize that it is critical for the scientific community to understand how to train competitive VLMs without doing so. \textbf{Molmo} and \textbf{PixMo} take an important step towards this understanding.


\nocite{Cherti_2023_CVPR}

\nocite{alayrac2022flamingo}
\nocite{li2022blipbootstrappinglanguageimagepretraining}  
\nocite{Tsimpoukelli2021MultimodalFL}

\nocite{peng2023kosmos2groundingmultimodallarge}  
\nocite{Dai2023InstructBLIPTG} 
\nocite{liu2024llavanext}
\nocite{hong2024cogvlm2}
\nocite{Cha2024HoneybeeLP}
\nocite{Thapa2024DragonflyMZ}
\nocite{liu2024sphinxx}  
\nocite{Li2023OtterAM}  
\nocite{Zhang2024LLaMAAdapterEF}  
\nocite{McKinzie2024MM1MA}  
\nocite{Zhang2024MM15MA}  
\nocite{li2024covlm} 
\nocite{lu2024deepseek}
\nocite{wang2024cogvlm} 
\nocite{laurenccon2024building}  
\nocite{zhang2024internlm}  
\nocite{li2023otterhd} 
\nocite{chen2023pali} 
\nocite{fu2024vita}  
\nocite{Li2024MonkeyIR}
\nocite{Sun2024Emu2}
\nocite{Hu2024mPLUGDocOwl1U}
\nocite{Yao2024DenseCF}


\nocite{Pramanick2023JackOA}
\nocite{Zhang2024Ferretv2AI}  
\nocite{peng2023kosmos2groundingmultimodallarge} 
\nocite{rasheed2024glamm}
\nocite{xiao2024florence} 
\nocite{Yuan2024OspreyPU}

\nocite{Li2023MIMICITMI}
\nocite{laurenccon2024unlocking}  
\nocite{jiang2024mantis} 
\nocite{fang2024vila2vilaaugmentedvila}  

\nocite{yuan2024tinygptv} 
\nocite{luo2023cheap} 
\nocite{chen2024evlmefficientvisionlanguagemodel} 
\nocite{lin2024moe} 
\nocite{chu2023mobilevlmfaststrong} 
\nocite{yao2024minicpm}
\nocite{zhu2024minigpt4}

\vspace{5mm}
\appendix


\global\csname @topnum\endcsname 0

\section*{Appendix}
The appendix includes the following sections:
\begin{itemize} 
\itemsep0em 
    \item \S\ref{supp:model} - Model Details
    \item \S\ref{supp:training} - Training Details
    \item \S\ref{sec:evaluation_supplementary} - Evaluation Details
    \item \S\ref{supp:result_details} - Result Details
    \item \S\ref{supp:ablations} - Ablation Details
    \item \S\ref{supp:data} - Data Details
    \item \S\ref{supp:data_examples} - Dataset Examples
    \item \S\ref{supp:related_work} - Related Work
\end{itemize}

\section{Model Details}
\label{supp:model}
We present additional details about image encoding, hyperparameters, and implementation choices.

\subsection{Image Encoding}
\begin{figure}[t]
    \centering
    \includegraphics[width=\linewidth]{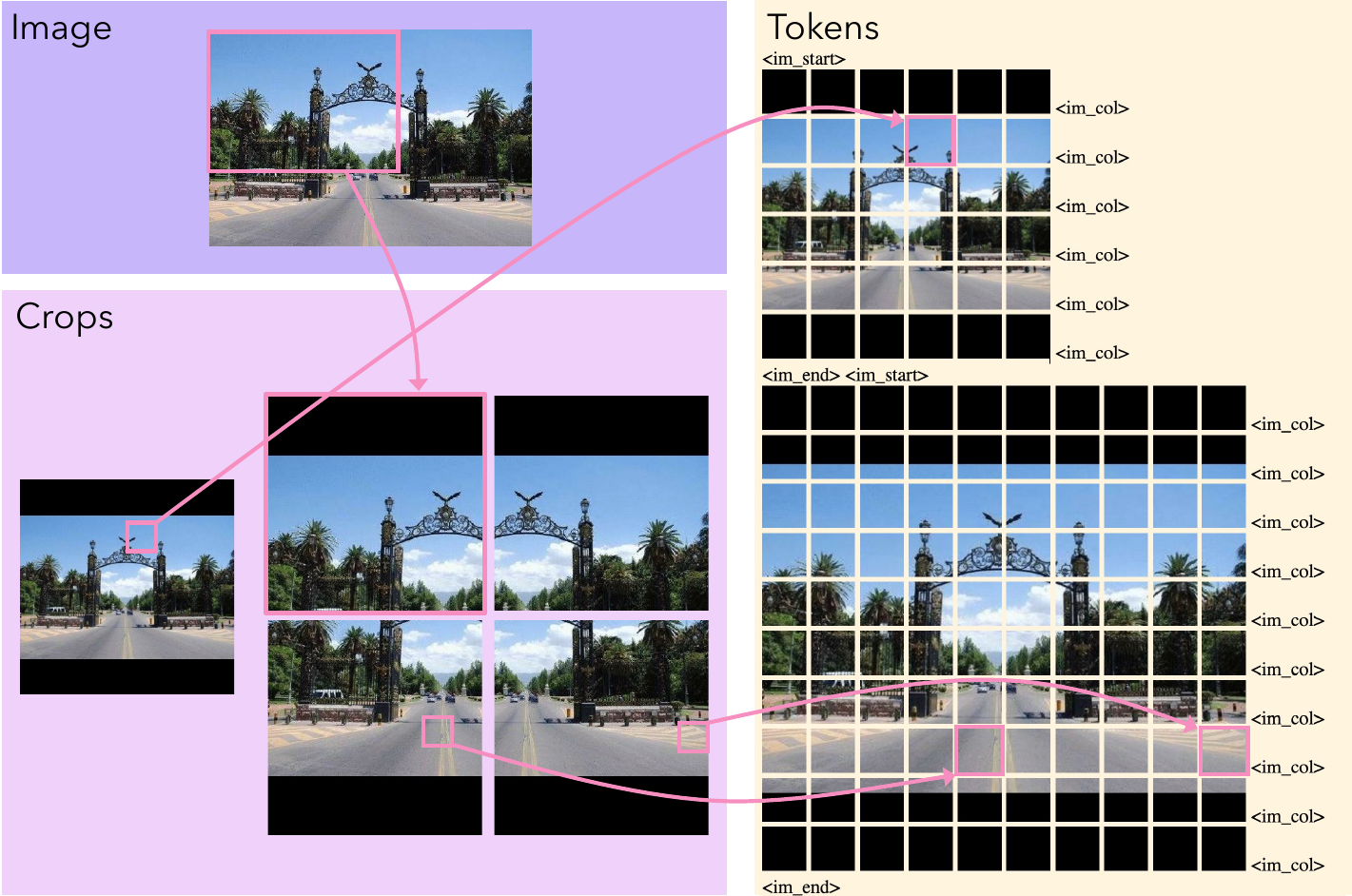}
    \caption{\textbf{Converting an image into tokens.} The image (top left) is turned into a single low-res and several overlapping high-res crops (bottom left). Padding (the black borders) is used so each crop is square and the aspect ratio of the image is preserved. The final token sequence for the image (right, arranged top-down left-to-right with line breaks for clarity) is built by extracting patch-level features from the crops, shown here using images of the patches, and special tokens. An image start and image end token are placed before/after the high-res and low-res patches, and column tokens are inserted after each row of patches. This example uses 4 high-res crops and extracts features from 36 (6$\x$6) patches per crop, in practice Molmo typically uses 12 high-res crops and extracts features from 144 (12$\x$12) patches per crop.}
    \label{fig:image-to-sequence}
\end{figure}

Our method of encoding images is shown in Figure~\ref{fig:image-to-sequence}. Cropping is done by first choosing a rectangular grid (\eg, a 2$\x$2, 3$\x$1, \etc) where each square of the grid matches the ViT's input size.
When using overlapping crops, these squares are moved closer together so that they overlap by a fixed margin (we use a margin of 4 patches or 56 pixels), which reduces the overall size of the grid.

Then the image is up-scaled to fit within that grid as well as possible while preserving its aspect ratio by making either its height or width the same size as the grid. The grid is chosen to require the least amount of up-scaling, and in the event of ties, to minimize its size.
We also set a maximum number of crops, and if the image cannot be covered by that many crops, the image will instead be down-scaled to fit the grid, and the grid is chosen to minimize the amount of down-scaling required while not exceeding the maximum number of crops. In either case, the re-scaled image is padded with black borders so that it exactly fits the grid, and then crops are extracted from this padded image. 
The low-resolution crop is built by resizing and padding the image so it matches image ViT's supported resolution.

Each crop is processed independently by the ViT and connector to get visual embeddings of each patch.  A learned embedding is added to the patch features from each crop (before the connector is applied) depending on whether that patch includes no padding, some padding, or is all padding, so the model can distinguish padding from images that naturally have black borders. These embeddings are arranged with special tokens as described in Section~\ref{sec:arch}, also shown in Figure~\ref{fig:image-to-sequence} right. For image/text inputs we encode the input image first, followed by any text.

\subsection{Hyper-Parameters}

\begin{table}[]
\newcommand\allmodels[1]{\multicolumn{4}{c}{\cellcolor{olive!5} #1}}
\footnotesize
 \begin{tabular}{c l c c c c}
  & & 1B-E & 7B-D & 7B-O & 72B-D  \\
\midrule
\multirow{8}{*}{\rotatebox[origin=c]{90}{{\footnotesize Image Encoder}}} 
 & Params & \allmodels{290m} \\ 
 & Dim & \allmodels{1024} \\
 & MLP Dim & \allmodels{4096} \\
 & Act. & \allmodels{GELU} \\
 & Heads & \allmodels{16} \\
 & KV Heads & \allmodels{16} \\
 & Layers & \allmodels{23} \\
 & Image Size & \allmodels{336$\x$336} \\
 & Patch Size & \allmodels{14} \\
 & Dropout & \allmodels{0.0} \\
 \midrule
\multirow{6}{*}{\rotatebox[origin=c]{90}{{\footnotesize V/L Connector}}} 
 & Params & 12m & 110m & 74m & 310m \\
 & Pool Size & \allmodels{2$\x$2} \\
 & Pool Dim & \allmodels{1024} \\
 & Pool Heads & \allmodels{16} \\
 & MLP Dim & 1024 & 37888 & 22016 & 59136 \\ 
 & Act. & \allmodels{SwiGLU} \\
 & Dropout & \allmodels{0.0} \\
 \midrule
\multirow{9}{*}{\rotatebox[origin=c]{90}{{\footnotesize LLM}}} 
 & Params & 1.2b (6.9b) & 7.6b & 7.3b & 72b \\
 & Embed & 50304 & 152064 & 100352 & 152064 \\
 & Dim & 2048 & 3584 & 4096 & 8192 \\
 & MLP Dim & 2048$\x$64 & 37888 & 22016 & 59136 \\
 & Act. & \allmodels{SwiGLU} \\
 & Heads & 16 & 28 & 32 & 80 \\
 & KV Heads & 16 & 4 & 32 & 8\\
 & Layers & 16 & 28 & 32 & 64 \\
 & Theta & 10k & 1m & 0.5m & 1m \\
 & Dropout & \allmodels{0.1} \\
 \midrule
\multirow{9}{*}{\rotatebox[origin=c]{90}{{\footnotesize Pre-Train}}} & Warmup ViT & \allmodels{2000}\\
 & Warmup Con. & \allmodels{200}\\
 & Warmup LLM & \allmodels{2000}\\
 & LR ViT & \allmodels{6e-6}\\
 & LR Con. & \allmodels{2e-4}\\
 & LR LLM & 2e-5 & 2e-5 & 2e-5 & 1e-5 \\
 & Cosine Decay & \allmodels{10\%} \\
 & Eps. & \allmodels{1e-6} \\ 
 & Betas & \allmodels{0.9, 0.95} \\ 
 & Batch Size & \allmodels{128} \\ 
 & Steps & \allmodels{22.3k} \\ 
 \midrule
 \multirow{8}{*}{\rotatebox[origin=c]{90}{{\footnotesize Fine-Tune}}} & Warmup ViT & \allmodels{200}\\
 & Warmup Con. & \allmodels{200}\\
 & Warmup LLM & \allmodels{200}\\
 & LR ViT & 5e-6 & 5e-6 & 5e-6 & 3e-6 \\
 & LR Con. & 5e-6 & 5e-6 & 5e-6 & 3e-6 \\
 & LR LLM & 2e-5 & 1e-5 & 1e-5 & 5e-6 \\
 & Cosine Decay & \allmodels{10\%} \\
 & Eps. & \allmodels{1e-6} \\ 
 & Betas & \allmodels{0.9, 0.95} \\ 
 & Batch Size & \allmodels{256} \\ 
 & Steps & 30k & 30k & 32k & 20k \\

\end{tabular}
    \caption{\textbf{Model and training hyper-parameters.} Molmo-1B-E has 1.2b active parameters, but 6.9b total. Its LLM MLP layers have 64 experts with 8 active at once.}
    \label{tab:hyperparameters}
\end{table}

Hyper-parameters for the Molmo models and the AdamW~\cite{kingma2015adam,loshchilov2019decoupled} optimizers are shown in Table~\ref{tab:hyperparameters}. The connector MLP uses the same intermediate dimension as the LLM, so its size depends on the LLM. The connector pooling layer and ViT architecture are the same between all models. All runs used a cosine learning rate schedule ending at 10\% of the peak learning rate~\cite{loshchilov2016sgdr}.

Learning rates are similar between the models, except we find it helpful to reduce the learning rate for Molmo-72B. We also find Molmo-72B learns faster than the other models and can therefore be trained for fewer steps. Molmo-7B-O was trained for slightly longer due to a minor configuration difference, but we do not think it affected performance.

\begin{figure}
    \centering
    \includegraphics[width=\linewidth]{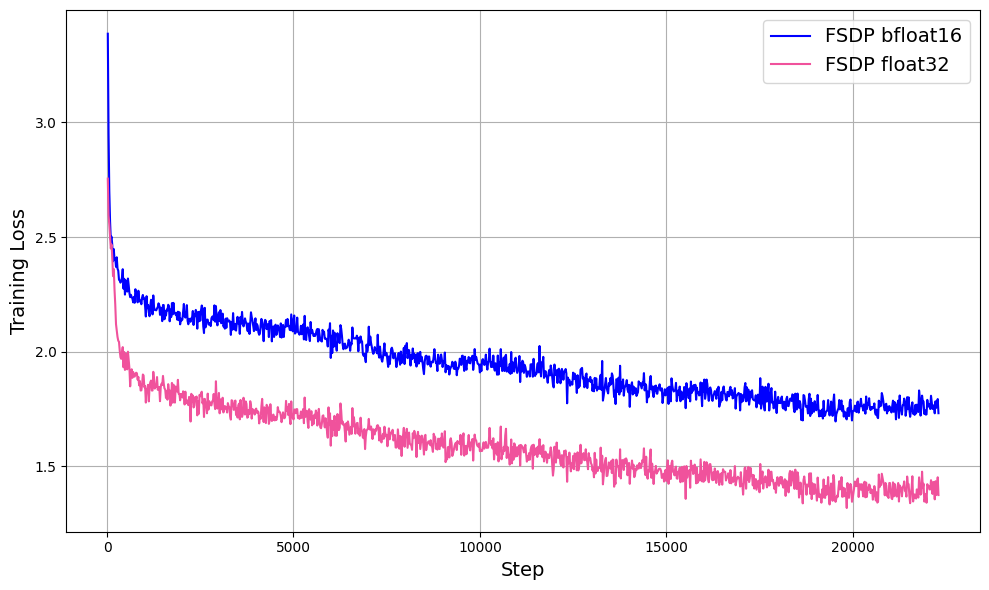}
    \caption{\textbf{Training loss curves} for Molmo-7B-D with model weights and gradient reduction in bfloat16 (\textcolor{blue}{blue}) and float32 (\textcolor{molmocolor}{pink}). Float32 is our default configuration.}
    \label{fig:fsdp-precision}
\end{figure}

\subsection{Implementation}
Our implementation uses PyTorch with Fully Sharded Data Parallel (FSDP)~\cite{zhao2023pytorch} based on the OLMo codebase~\cite{OLMo}.
We do not use FlashAttention~\cite{dao2022flashattention,dao2024flashattention2} since it does not support the more complex masks that are required for multi-annotated images, but we find using PyTorch's Scaled Dot Product Attention (SDPA) achieves close to the same speed.

To improve throughput, we utilize PyTorch's Automatic Mixed Precision (AMP) module\footnote{\url{https://pytorch.org/docs/stable/amp.html}}, which enables most operations to run in half-precision with bfloat16 numbers. However, as shown in Figure~\ref{fig:fsdp-precision}, keeping model weights and performing gradient reduction in half-precision degrades training loss, so these are retained in full precision. Additionally, computations for layer normalization~\cite{ba2016layer} and Rotary Position Embedding (RoPE)~\cite{su2024roformer} are explicitly carried out in full precision.

When computing gradients with FSDP, each GPU computes a gradient on a small mini-batch of examples, after which gradients are averaged across all devices. We always compute the per-device gradient by dividing the total loss on that device by the \textit{average} number of loss-tokens across all devices, not the number of loss tokens on that particular device. This avoids a subtle bias that effectively up-weights examples with a small number of loss tokens (\eg, with short responses) since those examples tend to be paired with a smaller divisor if using the device-local number of loss tokens. 
Using the average number of loss tokens across all devices largely resolves this issue since our global batches are much larger than the device-local batches.
This issue has been discussed in other places\footnote{\url{https://unsloth.ai/blog/gradient}}~\cite{hermans2017accumulated} and is known to have affected many codebases.\footnote{\url{https://github.com/huggingface/trl/issues/2175}}
We observe that captioning performance can drop by 0.5-1 points without this fix.

During fine-tuning, mixing is done within each batch so batches contain examples from a variety of tasks. We set a maximum sequence length of 2304 for both pre-training and fine-tuning, and truncate examples longer than that (in practice, truncation only happens for certain synthetic datasets like DVQA~\cite{kafle2018dvqa} which contains many annotations per image, or for the occasional outlier example in other datasets).

We find training to be stable, without loss spikes or NaNs, likely in part because we use pre-trained models.

\section{Training Details}
\label{supp:training}
Here we discuss the training mixture and how tasks are formatted during pre-training and fine-tuning.

\subsection{Pre-Training Task Details}
During pre-training, we train on each image paired with its caption and one of its audio transcripts. For images with multiple transcripts, we select one randomly each epoch. We use multi-annotation training (see Section~\ref{sec:evaluation_supplementary}) to train on both the caption and the transcript jointly.

We prompt the model with either ``\texttt{long\_caption:}" or ``\texttt{transcript:}" for captions and transcript respectively (a natural language prompt is used instead during instruction fine-tuning).
We also add a length hint: an integer providing a noisy hint as to the correct output length. This hint is computed as the length of the transcript/caption in characters, plus a noise factor drawn from a random normal with a standard deviation of 25. The hint is then divided by 15 and rounded down to keep the hint in roughly the range of 0 to 100. This noise is added so that the length functions more like a guideline than a hard constraint, leaving the model some flexibility to adjust the caption as appropriate for the image. For example, even with a long length hint, its preferable that a caption for a very plain image be short instead of becoming repetitive or inane due to lack of content to describe.

We add the hint to the prompt 90\% of the time, for example: ``\texttt{long\_caption\_83:}" for a length hint of 83, and 10\% of the time no length hint is used to maintain the ability to output a default caption.

\begin{figure}
    \centering
    \includegraphics[width=\linewidth]{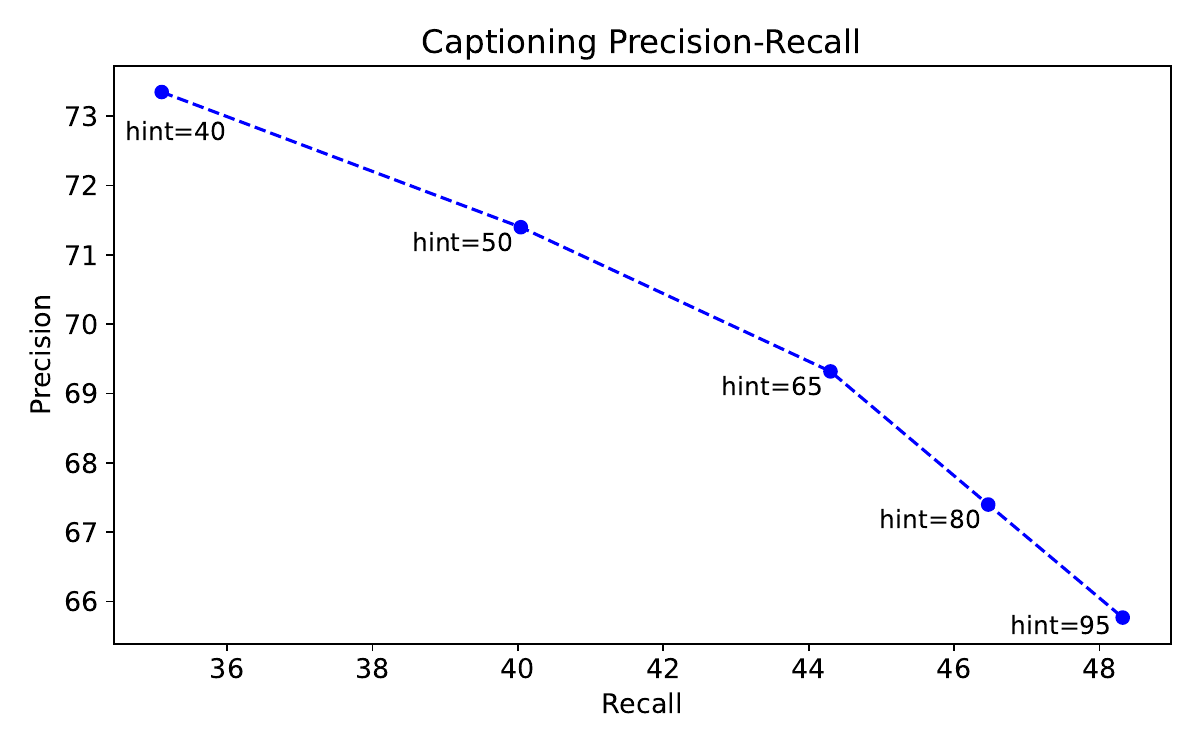}
    \caption{\textbf{Captioning precision and recall with different length hints} for Molmo-7B-D after pre-training. A short hint reduces recall since the model describes fewer things, but can boost accuracy since the description tends to focus on the more salient, easier-to-understand parts of the image.}
    \label{fig:length-conditioning}
\end{figure}

Adjusting the length hint allows a trade off between precision and recall in captioning, see Figure~\ref{fig:length-conditioning} (see Section~\ref{sec:evaluation_supplementary} for captioning metric details). 
In all of these settings, the average caption length when using a length hint is within 10 characters of the expected length, showing the models follow the length hints well.
For our ablations, we report scores with a length hint of 65, which performs similarly or slightly better than using no length hint.

Preliminary experiments with mixing in other sources of captions (COCO Captions~\cite{chen2015microsoft}, Localized Narratives~\cite{PontTuset_eccv2020}, or captions derived from Visual Genome annotations~\cite{Krishna2016VisualGC} ) did not improve scores on our captioning metric, so we use PixMo-Cap alone for pre-training.

\subsection{Fine-Tuning Task Details}
\label{supp:fine_tuning_details}
\begin{table}[]
\footnotesize
\newcommand\band{\rowcolors{1}{blue}{blue}}
\renewcommand{\indent}{\hspace{0.2cm}}
    \centering
    \begin{tabular}{l c c c c c }
         name & rate & images & anno. & tokens & avg.\hspace{0.03cm}crops \\ \midrule
\baseline{\textbf{PixMo (Annotated)}} & \baseline{38.1} & \baseline{1m} & \baseline{3.3m} & \baseline{350m} & \baseline{10.6} \\
\indent Points & 28.8 & 220k & 2.3m & 160m & 10.1 \\
\indent AskModelAnything & 3.8 & 71k & 160k & 17m & 10.0 \\
\indent Cap & 3.2 & 712k & 712k & 160m & 10.9 \\
\indent PointQA & 2.4 & 14k & 76k & 11m & 11.0 \\
\baseline{\textbf{PixMo (Synthetic)}} & \baseline{31.6} & \baseline{1.3m} & \baseline{3.3m} & \baseline{120m} & \baseline{11.0} \\
\indent Count & 6.2 & 36k & 37k & 3m & 11.9 \\
\indent CapQA & 5.7 & 160k & 210k & 38m & 10.8 \\
\indent Clocks & 5.3 & 800k & 800k & 20m & 10.5 \\
\indent Docs-Charts & 5.2 & 120k & 1.1m & 34m & 12.8 \\
\indent Docs-Other & 4.0 & 71k & 610k & 15m & 12.8 \\
\indent Docs-Tables & 3.3 & 47k & 420k & 12m & 12.1 \\
\indent Docs-Diagrams & 1.9 & 16k & 140k & 3.6m & 12.5 \\
\baseline{\textbf{Academic}} & \baseline{30.3} & \baseline{880k} & \baseline{25m} & \baseline{1b} & \baseline{8.0} \\
\indent TallyQA & 3.9 & 130k & 250k & 4.6m & 6.1 \\
\indent VQA v2.0 & 3.1 & 83k & 440k & 7.9m & 6.7 \\
\indent AndroidControl & 2.9 & 74k & 300k & 13m & 11.0 \\
\indent A-OKVQA & 2.8 & 17k & 17k & 380k & 6.8 \\
\indent DocVQA & 2.1 & 10k & 39k & 1m & 12.9 \\
\indent TextVQA & 2.0 & 22k & 35k & 700k & 12.7 \\
\indent ChartQA & 1.8 & 18k & 28k & 850k & 9.3 \\
\indent ST-VQA & 1.7 & 18k & 25k & 530k & 4.6 \\
\indent InfographicVQA & 1.6 & 4.4k & 24k & 670k & 12.0 \\
\indent TabWMP & 1.6 & 23k & 23k & 930k & 2.3 \\
\indent PlotQA & 1.5 & 160k & 20m & 930m & 12.4 \\
\indent AI2D & 1.3 & 6.2k & 15k & 630k & 6.4 \\
\indent DVQA & 1.1 & 200k & 2.3m & 51m & 5.0 \\
\indent FigureQA & 1.1 & 100k & 1.3m & 25m & 6.1 \\
\indent OK-VQA & 1.0 & 9k & 9k & 180k & 6.8 \\
\indent ScienceQA & 0.8 & 5k & 6.2k & 460k & 4.3 \\
    \end{tabular}
    \caption{\textbf{Full list of instruction fine-tuning tasks}. Columns show the sampling rate, the total number of images and annotations (\ie, the number of question/answer pairs), the number of text tokens using the Qwen2 tokenizer, and the average number of crops. The number of crops per an image can be at most 13 (one low-res, and 12 high-res), but can be lower for datasets with smaller images.
    Shaded rows show the total counts for all datasets in the category.}
    \label{tab:fine-tuning-tasks}
\end{table}

Table~\ref{tab:fine-tuning-tasks} shows a list of our fine-tuning tasks. We only train on the train sets. Formatting and task-specific details are described below.

\paragraph{Multiple choice questions.} For multiple choice questions in academic datasets (AI2D, A-OKVQA, ScienceQA), we append ``\texttt{Choices:}", a newline, and then new-line separated options with capital letter answer labels. The model predicts the answer label only. 
Note some multiple-choice questions appear in other, more diverse formats in PixMo-CapQA and PixMo-AskModelAnything.

\paragraph{Multiple answers.} For datasets with multiple answers per question (\eg, VQA v2.0), we only use the most common answer for training. If there are multiple answers that are equally common, we randomly select from among them each epoch.

\paragraph{Pointing.}
Pointing uses an HTML-like format. (x,y) coordinates are scaled to 0-100. For a single point, the format is:
\\ \\
\noindent \texttt{<point x="10.0" y="10.0" alt="alt text">Inline text</point>}
\\ \\
For multiple points the format is:
\\ \\
\noindent \texttt{<points x1="10.0" y1="10.0" x2="20.0" y1="20.0" ... alt="alt text">Inline text</points>}
\\ \\
Numbering the points makes counting easier because the total count is always the number of the last point.

When interacting with users, we generally replace the point(s) text with the inline text, and show the image with the points using the alt text as hover text. For pointing and counting, the inline and alt text are both the name of what is being pointed at. For pointing-as-an-explanation these fields can be different.

\paragraph{PixMo-Points.}
Counting or pointing with a very large number of objects can lead to very long sequence lengths. To avoid memory errors we do not train on data with more than 40 counts; we expect to remove this limitation in future iterations of Molmo.

\paragraph{PixMo-AskModelAnything.} ``How many" questions are common in PixMo-AskModelAnything, but are not accompanied with pointing data. We observe that this can lead to the model failing to point when asked counting questions. To resolve this, we heuristically detect such questions and prefix them with an instruction to not point (\eg, ``\texttt{Answer without points.}"), randomly selected from a pool of 20 such instructions.

\paragraph{AI2D.} 
AI2 Diagrams requires labeling regions of the images with letters, and then training the model to answer questions by predicting the correct region by returning its letter. Evaluations in the literature have been mixed between labeling the regions with opaque boxes (\eg, \cite{llama3,gpt4omini}) and transparent boxes (\eg, \cite{tong2024cambrian,liu2024llavanext,beyer2024paligemma}). 
We train in both settings and present our main results with transparent boxes. Results with opaque boxes are in Section~\ref{supp:result_details}. 

For AI2D questions where the answers are just letters, we list the multiple-choice options without a letter prefix.

AI2D does not have a validation set, so we built our own custom validation set by separating out 384 images (with roughly 2000 questions-answer pairs). None of our models are trained on this set.

\paragraph{ChartQA.} The ChartQA train set contains many synthetic questions (21k synthetic \vs 7k non-synthetic), which we observe can be noisy and lower quality. 
To reduce the weight of these examples we re-weight ChartQA so the total weight of the synthetic and non-synthetic examples are equal. This also means the training data better matches the validation and test data which are evenly split between synthetic and non-synthetic questions.

\paragraph{A-OKVQA.} We train on the multiple choice questions and, for questions not marked as difficult direct answer, also use them as direct answer questions by not using the answer options. We use different style tags for direct answer and multiple choice versions of the questions.

\paragraph{TabWMP.} For TabWMP we treat the task as short answer and do not show the multiple-choice options.

\paragraph{AndroidControl.} 
We train on four input-output configurations: low-level instruction to action, high-level goal to action, low-level and high-level inputs to action, and high-level goal to action with chain-of-thought reasoning. Only the instruction and screenshot are provided to the model as input; accessibility trees, action history, and a prompt with details like available actions are omitted. Target actions are represented as text output strings and (x,y) coordinates are scaled to 0-100 just as with regular pointing.

\subsection{Training Time}
\begin{table}[]
    \footnotesize
    \centering
    \begin{tabular}{@{}l c c c c c c c@{}}
         & \multicolumn{3}{c}{pre-train} & & \multicolumn{3}{c}{fine-tune} \\
         & GPUs & time & GPU hr. & & GPUs & time & GPU hr. \\ \hline
         1B-E & 8 & 33.3 & 264 & & 64 & 13.3 & 850 \\  
         7B-D & 64 & 8.6 & 550 & & 128 & 11.2 & 1.4k \\
         7B-O & 64 & 8.9 & 570 & & 128 & 13.5 & 1.7k \\
         72B & 128 & 33.3 & 4.2k & & 256 & 32.4 & 8.3k
    \end{tabular}
    \caption{Training times for the Molmo models using H100 GPUs.}
    \label{tab:training_time}
\end{table}
Training time and number of GPUs used are shown in Table~\ref{tab:training_time}. All models were trained with H100 GPUs with infiniband connectivity.

\section{Evaluation Details}
\label{sec:evaluation_supplementary}

\paragraph{Captioning metric (cap $F_1$).}
We measure captioning quality, relative to an evaluation set of 1500\footnote{A few evaluations were done with a super-set of 2730 images. We do not expect this to have affected results significantly since the 1500 were a random subset of the 2730.} images, using the harmonic mean of captioning \emph{precision} and \emph{recall}, \ie the $F_1$ score. 
The evaluation set was gathered through a similar protocol as PixMo-Cap (selecting a small number of images matching a diverse set of categories), but the images were selected manually and are disjoint from images in PixMo-Cap.
Each evaluation image has up to six audio transcripts associated with it. To define the precision and recall of a caption for an image, let $g$ be the generated caption and $T$ be the set of ground-truth transcripts for the image. We prompt GPT-4o to enumerate a list of all distinct atomic statements contained in $g$ and, separately, the transcripts in $T$. We then prompt GPT-4o to match each item in the list of atomic statements from $g$ to items in the list of atomic statements from $T$. To compute recall, we consider matches as true positives and unmatched items from $T$'s list as false negatives. To compute precision, we prompt GPT-4o with the raw transcripts and the list of statements from $g$ and ask it to say if each statement is \emph{consistent} (a true positive) or \emph{inconsistent} (a false positive) with the transcripts. (We avoid using the atomic statements from $T$ when computing precision because it's a potentially noisy processing step that is not necessary.) We average precision and recall over all images in the evaluation set and compute the $F_1$ score of the averaged precision and recall values to produce our final summary metric: cap $F_1$.

While this metric is imperfect (\eg, GPT-4o makes mistakes, the transcripts do not contain all true statements about the image, \etc.) we found that improvements to cap $F_1$ corresponded to improvements in our subjective impressions of caption quality, and thus it was a useful internal metric for guiding model and data design. Most of our model design and pre-training data decisions were based on improving captioning quality, see Section~\ref{supp:result_details} and Figure~\ref{fig:capf1_11avg_corr} for more discussion.

\paragraph{Human evaluation.}
We defined 10 question categories and crowd sourced image-question pairs, using the same workers as for other annotation tasks. This resulted in the following categories and image-question pair counts:
\begin{itemize}[leftmargin=5mm,topsep=1mm]
    \item Output formatting: 1525
    \item Fine-grained detailed QA: 1510
    \item General: 1504
    \item Documents: 1499
    \item Captioning: 1493
    \item Counting: 1490
    \item Homework: 1489
    \item Charts: 1473
    \item Named entities: 1448
    \item Creativity: 1420
\end{itemize}
We performed two human evaluations, the first for a large set of models presented in our main results Table~\ref{tab:benchmark_results} and a second for a smaller set of ablation models in Table~\ref{tab:ablation_elo}. To collect feedback, we presented an annotator with an image, a question, and the output of ``model A'' and ``model B'', without revealing the model identities. The annotator had five options: \emph{tie (both bad)}, \emph{tie (both good)}, \emph{model A is better}, \emph{model B is better}, or \emph{I don't know}. The last option was for cases where the annotator did not know the correct answer (\eg, a math problem they do not know how to solve). For the first study, we sampled image-question pairs randomly from the 10 categories, until we had collected \app450 feedback responses per model pair. For the second study, we used a refined methodology in which we first manually verified the quality of the question-image pairs, resulting in a fixed set of 500 questions (exactly 50 per category), all of which were used for each pair of models. After collecting the feedback, we removed the \emph{I don't know} responses and computed an Elo ranking using the Bradley-Terry model, following the methodology of Chatbot Arena~\cite{chiang2024chatbot}.

\paragraph{AndroidControl.}
To evaluate Molmo on AndroidControl, we provided only the task instruction (high-level or low-level) and the current screenshot as input. The results are reported on the in-domain data test set (IDD) and the metric is step-wise accuracy.


\section{Result Details}
\label{supp:result_details}

\paragraph{Chatbot Arena.} In Table~\ref{tab:lmsyschatbotarena} we report a summarized version of the vision leaderboard for queries in English from the Chatbot Arena~\cite{chiang2024chatbot}, an independent third-party VLM evaluation. Molmo-72B outperforms all the fully open and open weight models but lags behind some of the propriety VLMs. As noted in the paper, we did our own Elo evaluation (see Section~\ref{sec:evaluation}), in which Molmo-72B ranks higher (2nd place). The difference in rating likely stems from the types of questions evaluated. We do note our data includes many counting and image-description questions which are particular strengths of Molmo.

\begin{table}[t]
\centering
\tablestyle{2pt}{0.99}
\resizebox{\linewidth}{!}{%
\begin{tabular}{@{}lccc@{}}
model & score & 95\% CI & opennness \\
\shline
Gemini-Exp-114~\cite{team2024gemini} & 1278 & +28/-27 & API only \\
ChatGPT-4o-latest (20240903)~\cite{hurst2024gpt} & 1256 & +13/-13 & API only \\
Gemini-1.5-Pro-002~\cite{team2024gemini} & 1220 & +15/-14 & API only \\
Gemini-1.5-Flash-002~\cite{team2024gemini} & 1219 & +15/-17 & API only \\
GPT-4o-2024-05-13~\cite{hurst2024gpt} & 1213 & +9/-9 & API only \\
Claude 3.5 Sonnet (20240620)~\cite{anthropic2024claude} & 1187 & +9/-7 & API only \\
Claude 3.5 Sonnet (20241022)~\cite{anthropic2024claude} & 1184 & +15/-15 & API only \\
Gemini-1.5-Pro-001~\cite{team2024gemini} & 1158 & +9/-8 & API only \\
GPT-4-Turbo-2024-04-09~\cite{gpt4} & 1157 & +7/-10 & API only \\
Gemini-1.5-Flash-8B-Exp-0827~\cite{team2024gemini} & 1137 & +15/-13 & API only \\
GPT-4o-2024-08-06~\cite{hurst2024gpt} & 1131 & +18/-20 & API only \\
Gemini-1.5-Flash-8B-001~\cite{team2024gemini} & 1133 & +10/-15 & API only \\
GPT-4o-mini-2024-07-18~\cite{gpt4omini} & 1124 & +7/-9 & API only \\
\baseline{Molmo-72B} & 1115 & +18/-17 & Fully Open \\
Qwen2-VL-72B~\cite{wang2024qwen2} & 1113 & +15/-17 & Open Weight \\
InternVL2-26B~\cite{internvl2} & 1096 & +11/-10 & Open Weight \\
Pixtral-12B-2409~\cite{agrawal2024pixtral} & 1085 & +13/-14 & Open Weight \\
Llama-3.2V-90B-Instruct~\cite{llama3} & 1085 & +12/-14 & Open Weight \\
Gemini-1.5-Flash-001~\cite{team2024gemini} & 1087 & +8/-8 & API only \\
\baseline{Molmo-7B-D} & 1076 & +15/-18 & Fully Open \\
Yi-Vision~\cite{ai2024yi} & 1070 & +21/-26 & Distilled \\
Claude 3 Opus~\cite{anthropic2024claude} & 1073 & +6/-8 & API only \\
Qwen2-VL-7B~\cite{wang2024qwen2} & 1068 & +15/-14 & Open Weight \\
Llama-3.2V-11B-Instruct~\cite{llama3} & 1061 & +14/-14 & Open Weight
\end{tabular}
}%
\caption{\textbf{Chatbot Arena's vision leaderboard} for English queries. The table is up to date as of Nov. 13, 2024. We show up to 20 rows for clarity.}
\label{tab:lmsyschatbotarena}
\end{table}

\paragraph{Clock reading.}
PixMo-Clocks is a novel source of clock reading data, a data type that is missing from most VLM training data (for which data information is published; we cannot know for models without published data details, such as API-only models and many open-weight models such as Pixtral and Llama 3). PixMo-Clocks are entirely synthetic and show a variety of watch bodies and faces against plain backgrounds (see Figure~\ref{fig:qual_clocks} for examples).

We tested how well Molmo trained on this data performs on the in-the-wild clock reading benchmark introduced in~\cite{yang2022s}. The benchmark sources clock images from three different datasets, COCO~\cite{lin2014microsoft}, OpenImages~\cite{kuznetsova2020open} and Clock Movies, a newly collected dataset based on the film \textit{The Clock} (2010).\footnote{\url{https://www.imdb.com/title/tt2008009}} They are highly out-of-distribution relative to the PixMo-Clocks training data. We also benchmarked several API-only models and open-weight (+ open-data) models for comparison. We compare all of these VLMs against the model presented from~\cite{yang2022s} that is specialized at the single task of clock reading.

We used the same query for all the VLMs: ``What time is being shown? Please respond only with the time as hours and minutes in HH:MM format.'', and followed the official evaluation protocol.\footnote{\url{https://github.com/charigyang/itsabouttime}} Table~\ref{tab:clocks} highlights that all the VLMs including proprietary models struggle to read clocks, with the exception of Molmo; see the notable performance gaps between Molmo and other VLMs. Molmo-72B surprisingly underperforms Molmo-7B-D and MolmoE-1B. This might be partially due to that PixMo-Clocks accounts only for 5.3\% of the fine-tuning data mixture and we trained Molmo-72B for fewer steps than the others. Augmenting PixMo-Clocks with real-world clock images could potentially increase performance, closing the gap between Molmo and the specialized clock reading model.

Despite training on synthetic data, we qualitatively observe the clocking-reading capabilities generalize effectively to more complex questions and to captioning. An example is in Figure~\ref{fig:data} lower right of the main paper.

\begin{table}[]
\footnotesize
\centering
\begin{tabular}{@{}l c c c@{}}
  model & acc. & hour acc. & min. acc.\\
  \midrule
  GPT-4o-0513~\cite{hurst2024gpt} & 2.7 & 14.2 & 8.6 \\
  Gemini 1.5 Pro~\cite{team2024gemini} & 0.9 & 11.6 & 5.1 \\
  Claude-3.5 Sonnet ~\cite{anthropic2024claude} & 6.6 & 22.3 & 17.5 \\
  \midrule
  PaliGemma-mix-3B~\cite{beyer2024paligemma} & 6.1 & 21.0 & 15.8 \\
  Phi3.5-Vision-4B~\cite{abdin2024phi} & 1.9 & 12.0 & 7.6 \\
  Qwen2-VL-72B~\cite{wang2024qwen2} & \underline{9.1} & \underline{24.9} & \underline{18.4} \\
  InternVL2-Llama-3-76B~\cite{internvl2} & 3.3 & 16.3 & 9.9 \\
  Pixtral-12B~\cite{agrawal2024pixtral} & 1.7 & 11.9 & 6.7 \\
  Llama-3.2V-90B-Instruct~\cite{llama3} & 3.4 & 17.9 & 10.1 \\
  \midrule
  LLaVA-1.5-13B~\cite{liu2023llava} & 0.8 & 11.6 & 5.7 \\
  xGen-MM-interleave-4B~\cite{xue2024xgen} & 2.0 & 11.9 & 8.0 \\
  Cambrian-1-34B~\cite{tong2024cambrian} & 1.8 & 11.1 & 7.2 \\
  LLaVA OneVision-72B~\cite{li2024llava} & 5.7 & 17.9 & 15.4 \\
  \midrule
  \textcolor{molmocolor}{MolmoE-1B} & 65.8 & 77.9 & 74.1 \\
  \textcolor{molmocolor}{Molmo-7B-O} & 64.2 & 76.3 & 73.8 \\
  \textcolor{molmocolor}{Molmo-7B-D} & \textbf{68.2} & \textbf{78.6} & \textbf{76.0} \\
  \textcolor{molmocolor}{Molmo-72B} & 65.6 & 77.1 & 73.7 \\
  \midrule
  Specialized single-task model~\cite{yang2022s} & 78.9 & 84.2 & 82.9 \\
\end{tabular}
\caption{\textbf{Clock reading benchmark results}. We report the averages of overall, hour and minute accuracies, each evaluated on three different test sets based on COCO, OpenImages and Clock Movies, respectively. \textbf{Bold} numbers represent the highest VLM scores while the best numbers, excluding Molmo, are \underline{underlined}. We categorize models into five groups: (first) API-only, (second) open-weight, (third) open-weight and open-data, (four) the Molmo family and (five) the specialized clock reading model.}
\label{tab:clocks}
\end{table}

\paragraph{Pointing.}
To evaluate the model's pointing performance, we constructed an evaluation set of 493 image and pointing question pairs. Each example was manually verified to ensure that either there is no target object or each target object instance is annotated with a single point and an accurate segmentation mask. The segmentation masks are generated by SAM~\cite{kirillov2023segment} using each ground-truth point as a prompt.

For cases with no target object, precision and recall are calculated as 1 if the model responds correctly (\eg outputs ``This isn't in the image.'') and 0 otherwise. When a target object is present, we first compute the pairwise distances between the predicted points and the ground-truth points to serve as the cost in the Jonker-Volgenant algorithm~\cite{Jonker1987ASA,crouse2016implementing}, which we then use to assign each predicted point to one of the ground-truth points. We then use the verified segmentation masks to determine if each predicted point with an assignment is a true positive or false positive. Specifically, we calculate precision as the fraction of predicted points located within the segmentation mask of their assigned ground-truth point, and recall as the fraction of segmentation masks covered by predicted points.

Table~\ref{tab:pointing} demonstrates Molmo's superior pointing capability. Similar to captioning and counting, pointing performance declines when the number of crops are unequal at training and test time.

\begin{table}[]
    \footnotesize
    \centering
    \begin{tabular}{l c c c}
         model & precision & recall & $F_{1}$ \\ \midrule
         \textcolor{molmocolor}{MolmoE-1B} & 73.0 & 72.9 & 72.2 \\
         \textcolor{molmocolor}{Molmo-7B-O} & 75.7 & 75.5 & 75.1 \\
         \textcolor{molmocolor}{Molmo-7B-D} & 75.0 & 74.6 & 74.3 \\
         \textcolor{molmocolor}{Molmo-7B-D} (36 crops) & 58.4 & 58.7 & 58.1 \\
         \textcolor{molmocolor}{Molmo-72B} & \textbf{75.8}  & \textbf{75.4} & \textbf{75.2} \\
    \end{tabular}
    \caption{\textbf{Pointing evaluation results}. Pointing can perform poorly when $\#$ of crops are unequal at training and test time.}
    \label{tab:pointing}
\end{table}

\begin{table}[h!]
    \footnotesize
    \centering
    \begin{tabular}{lccc}
    \multirow{2}{*}{\# crops train, test} & \multirow{2}{*}{CountBenchQA} & Pixmo-Count  & \multirow{2}{*}{11-avg} \\
    & & \textcolor{gray}{val} & \\
    \hline
    12, 36$^*$ & 87.7 & 73.9 & 75.8 \\
    \baseline{12, 36} & 88.5 & 85.2 & 76.9 \\
    36, 36 & 88.9 & 87.4 & 77.2 \\
    12 \textrightarrow{} 36, 36 & 88.9 & 87.4 & 77.2  \\
    \multicolumn{3}{c}{~}\\
    \end{tabular}
    \caption{\textbf{High-resolution fine-tuning results.} Result on counting datasets (CountBenchQA and PixMo-Count val set) and the overall average (11-avg) using different numbers of crops at train and test. Note: 12, 36$^*$ uses a higher number of crops (36 crops) at test time for counting datasets, which leads to a much worse accuracy. Our default setting (highlighted in \baseline{gray}) uses the same number of crops (12 crops) during training and inference for counting datasets. We experiment with fine-tuning the 12-crop model at higher resolution and evaluating with 36 crops (12~$\rightarrow$~36, 36). }
    \label{tab:high_res_ft}
\end{table}

\paragraph{High-resolution fine-tuning.}
In Table~\ref{tab:resolution} of the main paper, we showed that training the model with higher resolution (\ie, more image crops) yields slight improvements on the 11-avg metric (from 76.9 to 77.2 when increasing the number of crops used in training from 12 to 36). Rather than directly training at a higher resolution, we explore fine-tuning the model initially trained with 12 crops using a higher resolution. Specifically, we continue training the 12-crop model for 3000 additional steps (10\% of the fine-tuning steps) with 36 crops, roughly halving the learning rates of the vision encoder (lr=2e-6), connector (lr=2e-6), and language model (lr=5e-6). We keep the global batch size at 256 and use a warmup of 200 steps for all modules. 

Table~\ref{tab:high_res_ft} presents the results. Note that simply increasing the number of crops at inference time (first row) leads to degraded performance on counting tasks (88.5 \textrightarrow{} 87.7 for CountBenchQA and 85.2 \textrightarrow{} 73.9 for PixMo-Count). This suggests that a mismatch between training and testing resolutions adversely affects counting performance. As a result, our default  model (second row) uses the same number of crops (12 crops) for counting datasets.

After fine-tuning the model with higher resolution (fourth row), we observe that its counting performance can recover when evaluated with 36 crops, matching that of the model trained directly with 36 crops (third row), without sacrificing the overall 11-avg performance. This demonstrates that a brief period of high-resolution fine-tuning can effectively restore counting capabilities without affecting the average performance.

\paragraph{Text-only benchmarks.} PixMo consists exclusively of multimodal image-text data, without any text-only data. To investigate the potential impact of training solely on multimodal data on performance in text-only tasks, we report the results on common text benchmarks which assess a wide range of capabilities. We carefully follow the setup used by Llama 3 \cite{llama3} for each task, ensuring that we can reproduce their numbers within the reported confidence intervals.  As shown in Table \ref{table:text-only}, the Qwen2 language model employed in Molmo-7B-D appears to lose some knowledge across various tasks as a result of multimodal fine-tuning. 

We run a small experiment at the 7B scale to test whether adding text-only data from Tulu 3\footnote{\url{https://allenai.org/papers/tulu-3-report.pdf}}~\cite{wang2023far} to our fine-tuning data mixture can address this issue. We use two different ratios: the entire dataset and a version with 10\% down-sampling. Incorporating this text data enhances model performance on text-only tasks, particularly those involving mathematical reasoning and programming. Interestingly, down-sampling to 10\% of the data leads to better results on most text-only tasks and improves the average performance across the 11 multimodal academic benchmarks.

\begin{table*}[t!]
    \centering
    \tablestyle{2pt}{1.02}
    \begin{tabular}{@{}lccccccc@{}}
    model &
      \multicolumn{1}{l}{MMLU \cite{mmlu}} &
      \multicolumn{1}{l}{MMLU-Pro \cite{mmlu_pro}} &
      \multicolumn{1}{l}{GSM-8k \cite{gsm8k}} &
      \multicolumn{1}{l}{MATH \cite{math}} &
      \multicolumn{1}{l}{ARC-C \cite{arc_c}} &
      \multicolumn{1}{l}{HumanEval \cite{humaneval}} &
      \multicolumn{1}{l}{11-avg} \\
    \hline
    Qwen2-7B (language model)                           & \textbf{70.2}  & \textbf{42.1}  & \textbf{71.8}  & \textbf{40.3}  & \textbf{87.5} & 47.6 & -  \\
    \baseline{Molmo-7B-D}                          & 64.6 & 32.2 & 58.8 & 11.5 & 81.5 & 36.6 & 76.9 \\
    Molmo-7B-D + Tulu 3                  & 64.9 & 38.6 & 67.7 & 8.3 & 84.5 & 51.2 & 76.9 \\
    Molmo-7B-D + Tulu 3 $\times$ 0.1 down-sample & 65.4 & 37.3 & 71.2 & 27.5 & 84.9 & \textbf{55.5} & \textbf{77.1}
    \end{tabular}
\caption{\textbf{Text-only benchmark results.} 11-avg denotes the average performance on 11 academic benchmarks.}
\label{table:text-only}
\end{table*}

\paragraph{Human evaluation.}
\begin{figure}[t]\centering
\includegraphics[width=1.0\linewidth]{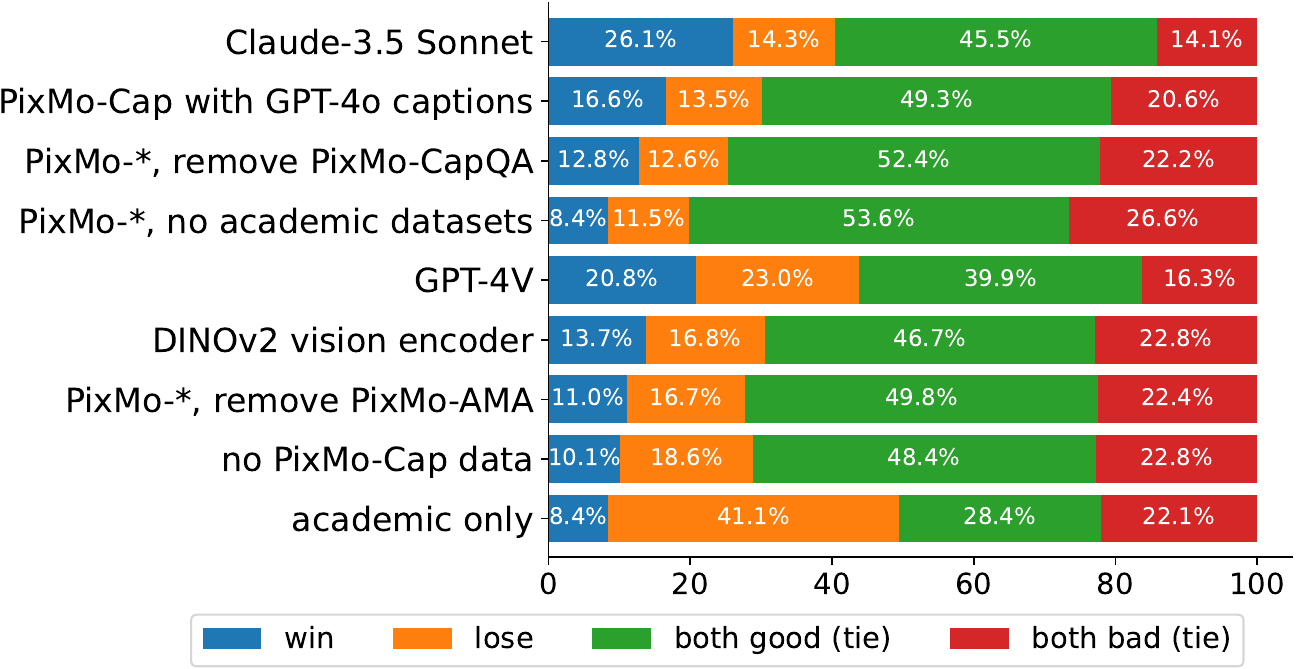}
\caption{\textbf{Human evaluation outcomes for matches between various models \vs \mbox{Molmo-7B-D}.} We expand upon the win rate (excluding ties) shown in Table~\ref{tab:ablation_elo} to report the full breakdown of wins, losses, ties (both good), and ties (both bad). We removed the \emph{I don't know} responses, which accounted for 2.9\% of all human feedback, before calculating the outcome rates.}
\label{fig:ablation_elo_rates}
\end{figure}
Table~\ref{tab:ablation_elo} of the main paper reports the win rates of several ablation and API-only models \vs Molmo-7B-D \emph{when ties are excluded} (a standard metric reported in the LMSYS Chatbot Arena). Ties make up a significant portion of the matches, so we report the full breakdown of match outcomes in Figure~\ref{fig:ablation_elo_rates} to better characterize the human evaluation. For example, when paired against Claude-3.5 Sonnet 45.5\% of matches resulted in a tie where both responses were good, 14.1\% in a tie where both responses were bad, Claude won 26.1\% of the time, and Molmo-7B-D won 14.3\% of the time. As a second example, compared to Molmo fine-tuned only on academic data the breakdown is 28.4\% (ties, both good), 22.1\% (ties, both bad), 8.4\% (it wins), 41.1\% (Molmo-7B-D wins).

\paragraph{AI2D with opaque boxes.} Table~\ref{tab:ai2d} shows result with and without opaque boxes on AI2D. The two options are described and discussed in Section~\ref{supp:fine_tuning_details}.

\begin{table}[]
    \tablestyle{2pt}{1.02}
    \centering
    \begin{tabular}{l c c}
         model & opaque & transparent \\ \midrule
         \textcolor{molmocolor}{MolmoE-1B} & 75.7 & 86.4 \\
         \textcolor{molmocolor}{Molmo-7B-O} & 79.8 & 90.7 \\
         \textcolor{molmocolor}{Molmo-7B-D} & 82.4 & 93.2 \\
         \textcolor{molmocolor}{Molmo-72B} & 86.4  & 96.3 \\
    \end{tabular}
    \caption{AI2D test scores with transparent and opaque boxes.}
    \label{tab:ai2d}
\end{table}

\paragraph{Cap $F_1$ and 11-avg correlation.}
\begin{figure}
    \centering
    \includegraphics[width=\linewidth]{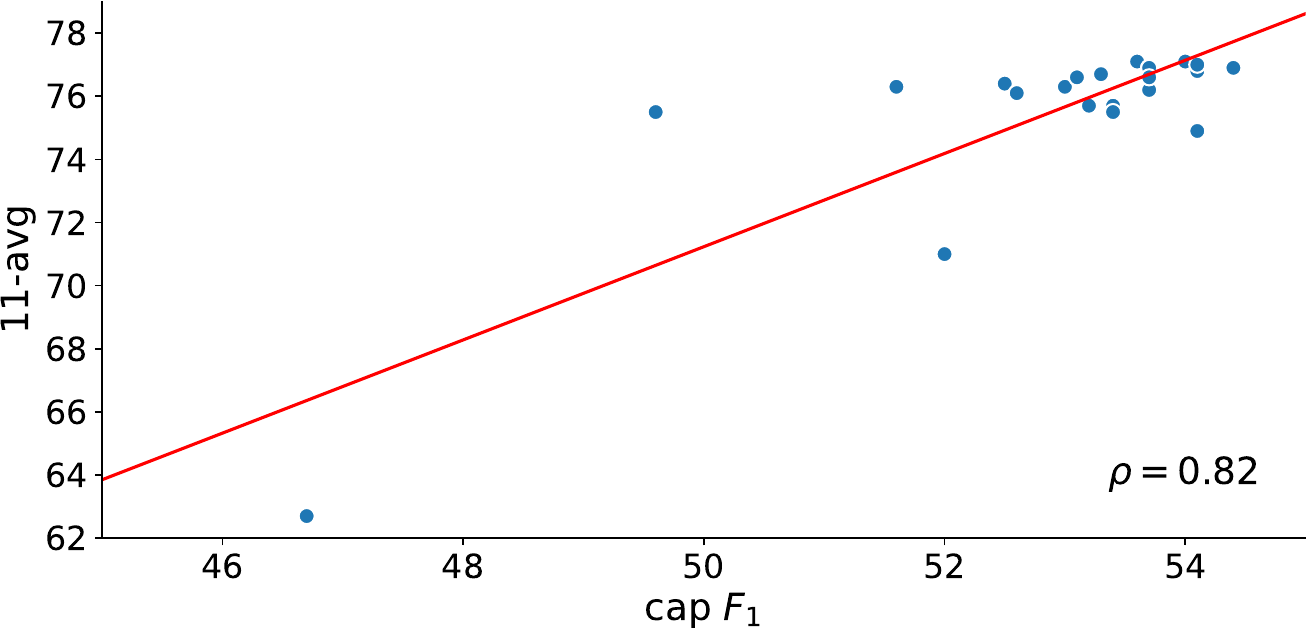}
    \caption{\textbf{Relationship between cap $F_1$ and 11-avg.} Our model development was driven by increasing cap $F_1$. Here, we show a scatter plot of cap $F_1$ \vs the 11 benchmark average (11-avg) from 22 ablation experiments, including all ablations that: (1) affect pre-training and (2) use PixMo-Cap. The Pearson correlation ($\rho$) is 0.82 and a least-squares regression line is shown in red.}
    \label{fig:capf1_11avg_corr}
\end{figure}
For the majority of the project we did not look at downstream tasks\footnote{We did a small number of sanity checks on VQA v2.0.} and instead made most modeling decisions to maximize our captioning metric. At the conclusion of the project, we used our ablation experiments to analyze the relationship between cap $F_1$ and the 11 benchmark average (11-avg), shown in Figure~\ref{fig:capf1_11avg_corr}. The scatter plot includes results from the 22 experiments that meet two conditions: (1) the experiment affects pre-training and (2) the experiment uses PixMo-Cap. We exclude a small number of experiments that use different pre-training data, \eg ShareGPT4o/v, because cap $F_1$ becomes an out-of-domain evaluation that is not directly comparable to in-domain results. We observe a strong correlation (Pearson $\rho=$ 0.82), suggesting that optimizing for dense captioning may be a reasonable proxy for a broad range of downstream tasks---though we have not established a causal relationship and this suggestion should be taken with that caveat in mind.

\paragraph{Leaderboards.}
We submitted Molmo-72B-D to several leaderboards. Molmo-72B-D achieves first on the VQA v2.0 leaderboard on the A-OKVQA leaderboard, achieves third on DocQA and InfoQA, behind QwenVL-72B, InternVL2-Pro.
\footnote{Results as of Nov. 21, 2024.}

\section{Ablations Details}
\label{supp:ablations}

\subsection{Discussion of Main Paper Ablations}
\paragraph{Vision encoder.} We adopted OpenAI's CLIP early on and used it for our main results and as the default in our ablations. Later, we evaluated the three alternative choices in Table~\ref{tab:vision_encoder}. All encoders are ViT-L/14 with 336$\x$336 pixel inputs, except for SigLIP which uses 384$\x$384 pixels. For MetaCLIP, we started with the weights of the 224$\x$224 model and resized the positional embeddings to 336$\x$336 before using it in Molmo. To equalize computation, we slightly reduced the maximum number of crops for SigLIP so that the average vision token count is similar for all models. Overall, the three encoders that were trained on web-scale noisy image-text data perform very similar to each other on both metrics. Of significant note, this includes MetaCLIP which is a \emph{fully} open model (data and weights) meaning that \emph{every model component and every bit of data in a Molmo model equipped with MetaCLIP and OLMo is open}. In retrospect, we should have used MetaCLIP as our default vision encoder, but we evaluated it too late in the process to retrain all Molmo models and ablations that were already based on OpenAI's CLIP vision encoder.

Also, surprisingly, when using the DINOv2 backbone---which is trained on images only (no text, no label supervision)---Molmo performs only slightly worse than the vision-language supervised vision encoders. DINOv2 also performs well in our user study (Table~\ref{tab:ablation_elo}), with a win-rate (excluding ties) of 45\% compared to our standard Molmo-7B-D configuration (\ie, Molmo-7B-D wins 55\% of all non-tie matches against its DINOv2-based variant).

\paragraph{Image resolution.} Table~\ref{tab:resolution} shows that using more crops (and thus higher image resolution) for training and testing generally improves results. We found that some tasks, like document-heavy ones, even benefit from using more crops in inference than the number used for training. However, for captioning and pointing (and thus counting), results degrade when the number of test crops does not equal the number of training crops. Therefore, for captioning and counting tasks we always force these values to be the same. As shown in Table~\ref{tab:high_res_ft} this awkward detail can be remediated by a small amount of high-resolution fine-tuning and then always using that same number of crops during inference for all tasks.

\paragraph{Dropout.} Dropout in the LLM generally improves both pre-training and fine-tuning (Table~\ref{tab:dropout}). We also find that our novel text-token-only dropout, in which dropout is only applied to the text tokens of the caption, not to the vision or prompt tokens, improves the captioning metric. We hypothesize that this restricted dropout encourages the model to rely more on the vision tokens, rather than guess based on the previous text tokens, when generating tokens which may reduce hallucinations.

\paragraph{Length conditioning.} Our captioning pre-training task includes a length hint. In Table~\ref{tab:len_cond} we ablate this design choice and find that it significantly impacts the captioning metric, but also improves the downstream tasks. Note that length conditioning only changes the pre-training task; the fact that it improves the downstream metrics indicates that captioning with length conditioning is a better pre-training task than just captioning.

\paragraph{PixMo-Cap scaling.} In Table~\ref{tab:scaling} we show the scaling effects of PixMo-Cap data by training with smaller fractions of the data in both pre-training and as part of the fine-tuning data mixture. Both metrics clearly improve as the amount of captioning data varies from none at all to the full set of 712k images. We also tested the model with no PixMo-Cap data in our user study (Table~\ref{tab:ablation_elo}), where it had a win-rate (excluding ties) of only 35\% compared to our standard \mbox{Molmo-7B-D} configuration. Removing PixMo-Cap data has a severe negative impact on its user preference score.

\paragraph{Pre-training data.} We consider different choices of pre-training data in Table~\ref{tab:stage1}. Pre-training VLMs, not just the vision encoder, with web-scale noisy image-text data is a popular data choice in contemporary methods (\eg,~\cite{beyer2024paligemma,llama3}). We test if this has any advantage using data from LAION~\cite{schuhmann2022laion5bopenlargescaledataset}. 
To do this we add a preliminary training stage that tunes the model for 50k steps and a batch size of 1024 on image/text pairs from LAION 2B. 
In this stage only the V/L connector is tuned, the LLM and image encoder are frozen. This pre-trained model is then trained on the dense captions and then our instruction tuning mixture as normal.
We find no improvement in metrics using this strategy, allowing us to keep the training pipeline simple. 

Another popular choice is to use ShareGPT4V/o~\cite{chen2023sharegpt4v}, which involves distilling from GPT-4 through captions. Using this data instead of PixMo-Cap performs worse on both metrics even when approximately controlling for the data scale (compare to 178k PixMo-Cap images in Table~\ref{tab:scaling}. In contrast, if we caption all PixMo-Cap images with GPT-4o and train on those captions, both metrics perform strongly. We think this is likely because PixMo-Cap has a more diverse image distribution, and due to captioning improvements in GPT-4o. Finally, we compare our default setting to either using only the raw audio transcripts or only using the LLM cleaned transcripts, both of which perform slightly worse than our default strategy of using both.

\paragraph{Supervised fine-tuning data.} We explore choices of fine-tuning data in Table~\ref{tab:stage2}. Using only academic data sets (specifically the ones in Fig.~\ref{fig:mixture}, but excluding AndroidControl) performs significantly worse than our full mixture (72.2\% \vs 76.8\%). The gap is primarily explained by PixMo-Docs, which improves results on document-heavy tasks, and the counting data from PixMo-Points and PixMo-Count. The other fine-tuning PixMo datasets have a small, and sometimes slightly negative, impact on the 11 benchmarks; they primarily add new skills to the model and improve user experience when chatting with it, as shown by user preference scores in Table~\ref{tab:ablation_elo}.

\paragraph{Counting.} In Table~\ref{tab:counting_special_tokens} we compare encoding points in plain-text as numbers between 0.0 and 100.0 with one significant digit of precision (our default) \vs adding 1000 special point tokens to the model's tokenizer, maintaining the same spatial precision. We find that using special point tokens performs substantially worse than the simple plain-text representation.

\begin{table*}[t]
\vspace{-.2em}
\centering
\subfloat[
\textbf{Vision encoder layers}. Concatenating features from multiple ViT layers (near the end and towards the middle) \vs using features only from each of those layers.
\label{tab:vision_enc_layers}
]{
\begin{minipage}{0.27\linewidth}{\begin{center}
\tablestyle{2pt}{1.02}
\begin{tabular}{lcc}
layers & cap $F_1$ & 11-avg \\
\hline
\baseline{3rd-to-last \& 10th-to-last} & \textbf{54.1} & \textbf{76.9} \\
only 3rd-to-last & 53.7 & 76.6 \\
only 10th-to-last & 52.5 & 76.3 \\
\end{tabular}
\end{center}}\end{minipage}
}
\hspace{2em}
\subfloat[
\textbf{Learning rate warmup}. The number of learning rate warmup steps used in caption pre-training for the three principal model components: ViT, connector, and LLM.
\label{tab:lr_warmup}
]{
\begin{minipage}{0.27\linewidth}{\begin{center}
\tablestyle{2pt}{1.02}
\begin{tabular}{lcc}
steps (ViT / con. / LLM) & cap $F_1$ & 11-avg \\
\hline
\baseline{2000 / 200 / 2000} & \textbf{54.1} & \textbf{76.9} \\
200 / 200 / 200 & 53.7 & \textbf{76.9} \\
\multicolumn{3}{c}{~}\\
\end{tabular}
\end{center}}\end{minipage}
}
\hspace{2em}
\subfloat[
\textbf{Grad norm}. Gradient normalization applied separately to the ViT, connector, and LLM parameters, globally to all parameters (row 2), or component-wise for pre-training and globally for fine-tuning (row 3). 
\label{tab:grad_norm}
]{
\begin{minipage}{0.27\linewidth}{\begin{center}
\tablestyle{2pt}{1.02}
\begin{tabular}{lcc}
grad norm & cap $F_1$ & 11-avg \\
\hline
\baseline{component-wise} & \textbf{54.1} & \textbf{76.9} \\
global & 53.6 & \textbf{76.9} \\
global, fine-tune only & \textbf{54.1} & \textbf{76.9} \\
\end{tabular}
\end{center}}\end{minipage}
}
\caption{\textbf{Additional model ablations.} Defaults are in \colorbox{baselinecolor}{gray}.}\label{tab:ablations_additional}\vspace{-4mm}
\end{table*}

\subsection{Additional Ablations}
Additional model ablations are presented in Table~\ref{tab:ablations_additional} for vision encoder layers, learning rate warmup, and gradient normalization. See table captions for more details.

\section{Data Details}
\label{supp:data}
\begin{figure}[t]\centering
\includegraphics[width=1.0\linewidth]{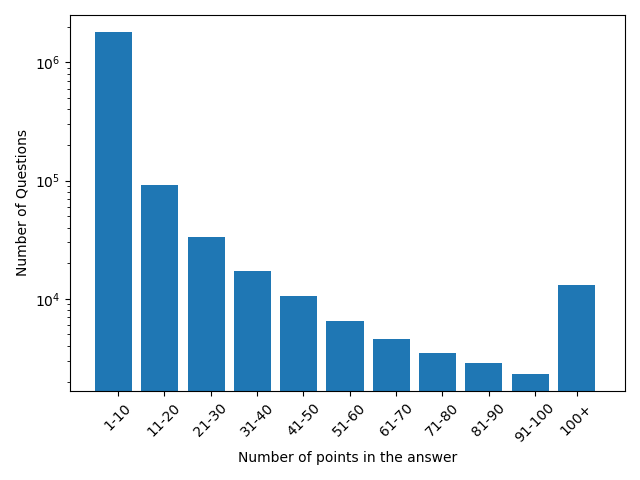}
\caption{\textbf{PixMo-Points distribution of counts.} We show the number of pointing questions (on a log scale) with answers in different ranges (\eg, 1 to 10, 11 to 20, \etc.).}\vspace{-4mm}
\label{fig:pixmo_point_count_answer_dist}
\end{figure}

\paragraph{PixMo-Points.}
The PixMo-Points dataset has a total of 229k unique images and a total of 1.98M referring expressions. It has an average of 8.7 distinct expressions per image with an average 5.5 points per expression, and an average of 47.7 total points per image. Additionally, there are 359k instances with no target object (no points). Figure~\ref{fig:pixmo_point_count_answer_dist} shows the distribution of number of points for expressions with non-zero points. 
PixMo-Points is a much larger and more diverse dataset than previous works such as gRefCOCO~\cite{GRES} (which contains a total of 20k images, 60k distinct instances, 278k expressions, of which 80k are multi-target and 32k are no-target expressions) and also much larger than RefCOCO, RefCOCOg and RefCOCO+~\cite{yu2016modeling}, each with about 86k, 142k and 141k unique referring expressions respectively and no multi-target references. 
Additionally, PixMo-Points focuses on referring to points and not segmentation masks, making it significantly more efficient to collect.

\paragraph{PixMo-Cap.}
We prompted our annotators with the following questions to answer in their spoken image descriptions.
\begin{enumerate}[leftmargin=5mm,topsep=1mm]
    \item \emph{What is the image at first glance?}
    \item \emph{What are the objects and their counts?}
    \item \emph{What does the text say?}
    \item \emph{What are the positions of the objects?}
    \item \emph{What subtle details are noticeable?}
    \item \emph{What is in the background?}
    \item \emph{What is the style and color?}
\end{enumerate}

\paragraph{PixMo-Docs.}
We developed a generation framework for synthesizing text- and figure-heavy images.
The core idea is to harness the coding capabilities of a text-only LLM to generate programs that render image data. 
These programs are then used as context for another LLM to construct instruction-tuning datasets.

Our framework supports seven programming languages/rendering libraries, including Matplotlib, Plotly, LaTeX, HTML, Vega-Lite, Mermaid, and Graphviz. 
Using these tools, we designed specialized pipelines to generate charts, tables, diagrams, and various types of documents.

The framework accepts text input to control the generation process. 
For instance, given the input ``restaurant menu'', the system selects the appropriate tools to generate relevant data. 
To diversify the final datasets, we use a comprehensive set of input queries. 
Additionally, we enhance data diversity by incorporating personas~\cite{ge2024scaling}, which control the content and style of the synthetic data. 
For example, when generating ``restaurant menu'' data with the persona ``A barbecue enthusiast known for their amazing grilled food at every Tennessee Vols game'', the framework produces a data point featuring a ``Southern fusion menu combining traditional BBQ with international flavors, presented on a wooden board background''.
This approach allows us to enrich the variety within each category of synthetic data.

We use Claude-3.5 Sonnet~\cite{anthropic2024claude} for code generation and GPT-4o-mini~\cite{gpt4omini} during the instruction-tuning data generation stage, prioritizing cost efficiency.




\begin{figure*}[!t]\centering
\includegraphics[width=\linewidth]{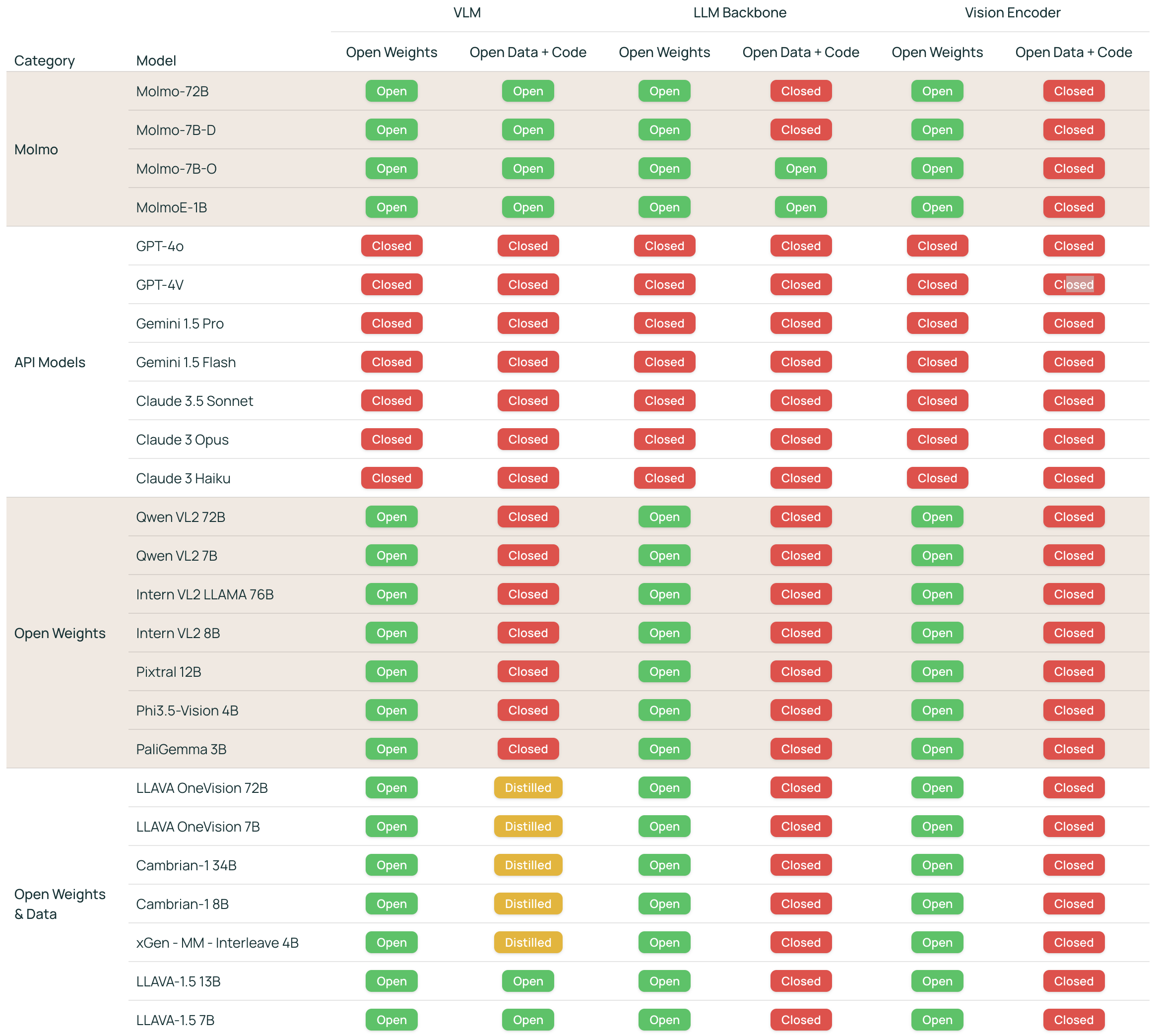}
\vspace{-3mm}
\caption{\textbf{VLM Openness Comparison.} We characterize the openness of VLMs based on two attributes (open weights, open data and code) across three model components (the VLM and its two pre-trained components, the LLM backbone and the vision encoder). In addition to open vs. closed, we use the "distilled" label to indicate that the data used to train the VLM includes images and text generated by a different, proprietary VLM, meaning that the model cannot be reproduced without a dependency on the proprietary VLM.
}
\label{fig:open}
\end{figure*}

\section{Dataset Examples}
\label{supp:data_examples}
We include \textbf{randomly selected} examples from the \textbf{\mbox{PixMo-$\star$}} datasets. Prompts are shown in bold, and points are shown with pink dots. 

\begin{figure*}
    \centering
    \includegraphics[width=\linewidth]{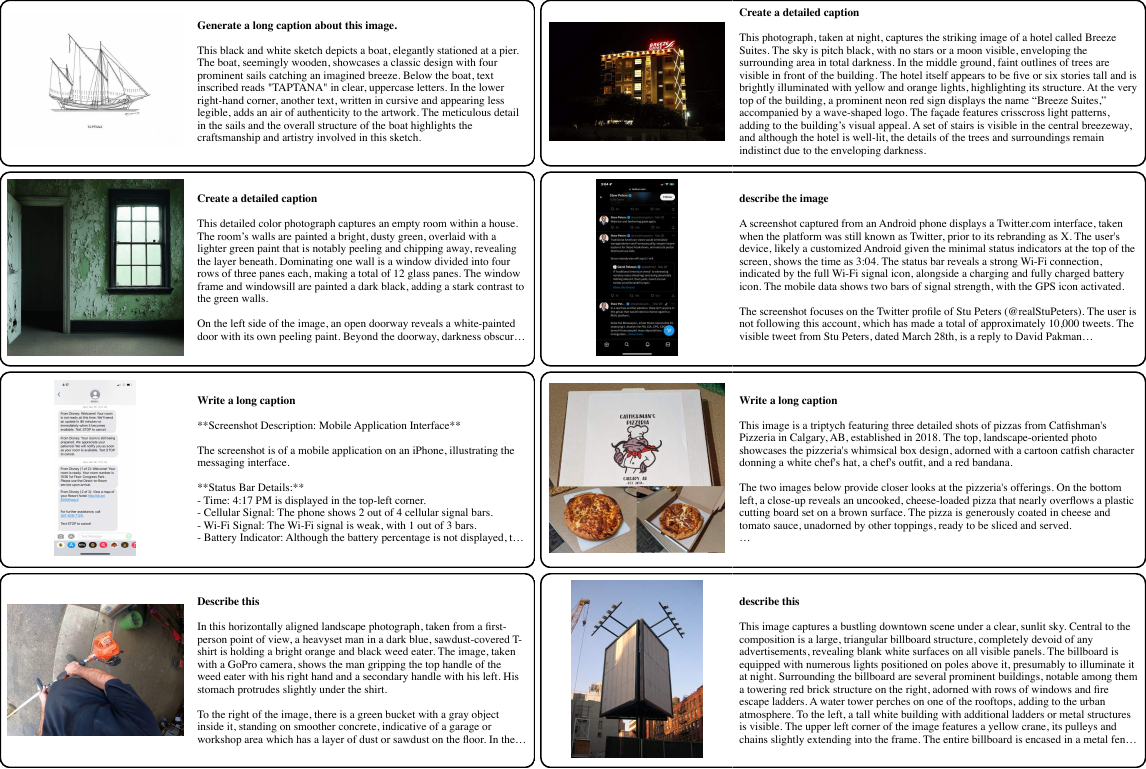}
    \caption{Randomly selected examples from \textbf{PixMo-Cap} with our prompt templates.}
    \label{fig:qual_cap}
\end{figure*}

\begin{figure*}
    \centering
    \includegraphics[width=\linewidth]{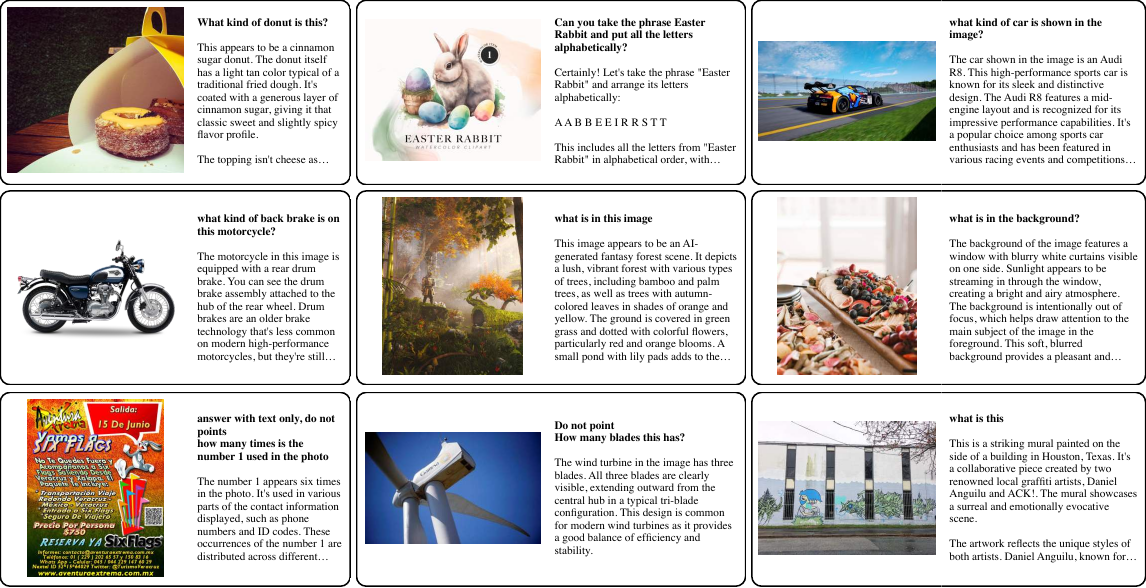}
    \caption{Randomly selected examples from \textbf{PixMo-AskModelAnything}.}
    \label{fig:qual_ask_model_anything}
\end{figure*}

\begin{figure*}
    \centering
    \includegraphics[width=\linewidth]{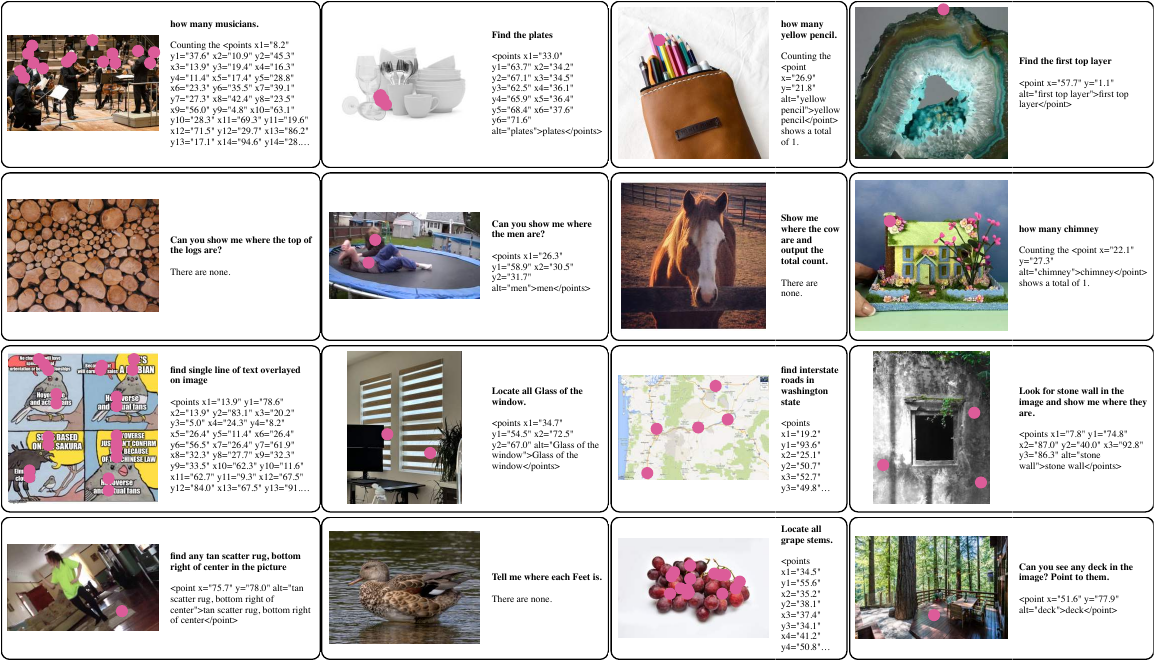}
    \caption{Randomly selected examples from \textbf{PixMo-Points}. Even when text has been cut off, all points still appear in the image.  Our templated prompts can be ungrammatical for some of these options, but we find they are still sufficient to let the model respond correctly to natural language instructions.}
    \label{fig:qual_points}
\end{figure*}

\begin{figure*}
    \centering
    \includegraphics[width=\linewidth]{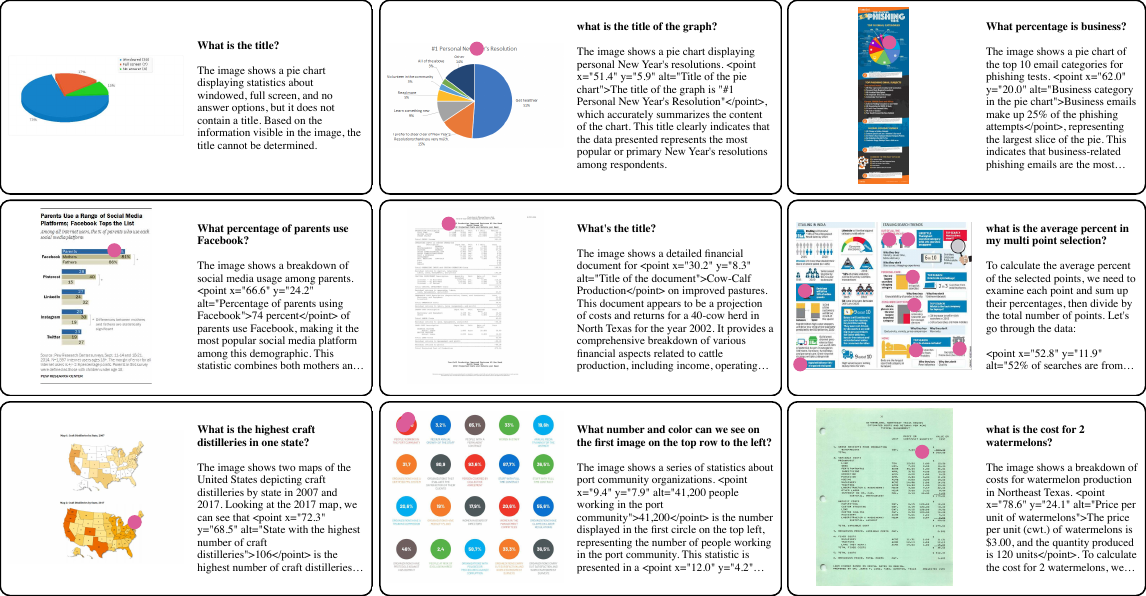}
    \caption{Randomly selected examples from the experimental \textbf{PixMo-Points} data that includes points with explanations.}
    \label{fig:qual_point_qa}
\end{figure*}

\begin{figure*}
    \centering
    \includegraphics[width=\linewidth]{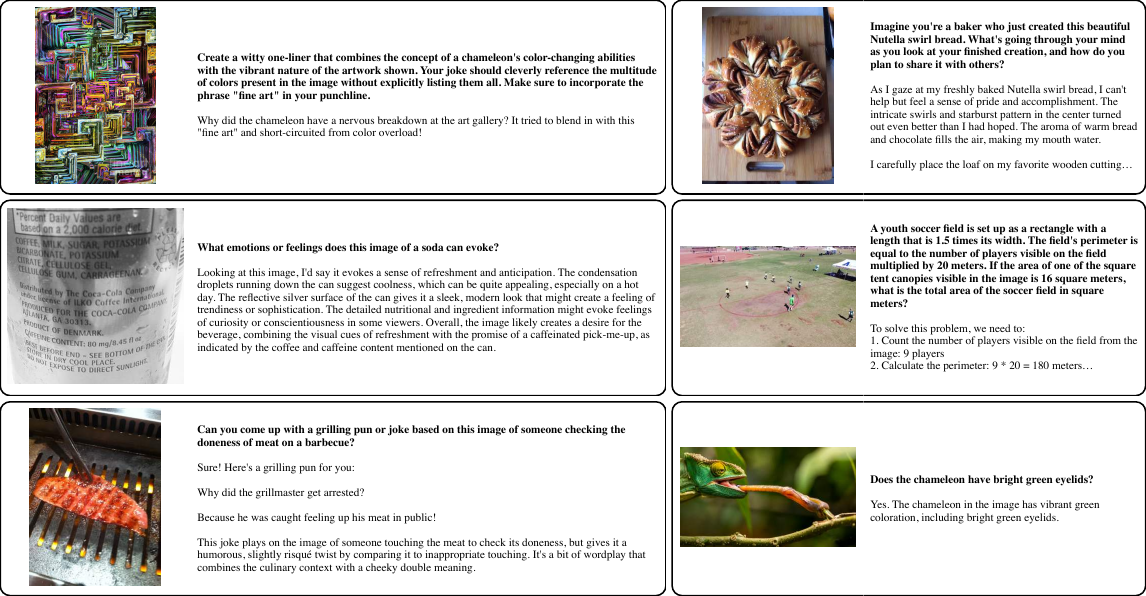}
    \caption{Randomly selected examples from the synthetic \textbf{PixMo-CapQA} data generated from PixMo-Cap.}
    \label{fig:qual_cap_qa}
\end{figure*}

\begin{figure*}
    \centering
    \includegraphics[width=\linewidth]{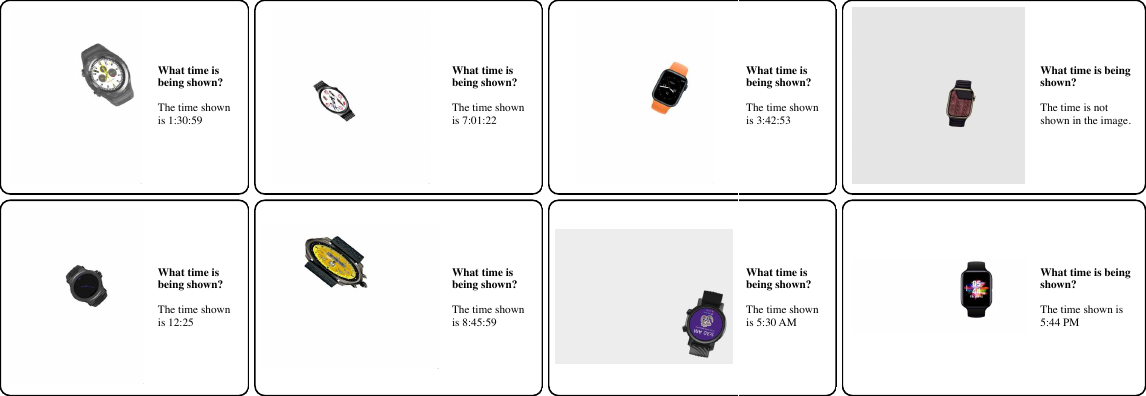}
    \caption{Randomly selected examples from the synthetic \textbf{PixMo-Clocks} data after our data augmentation.}
    \label{fig:qual_clocks}
\end{figure*}

\begin{figure*}
    \centering
    \includegraphics[width=\linewidth]{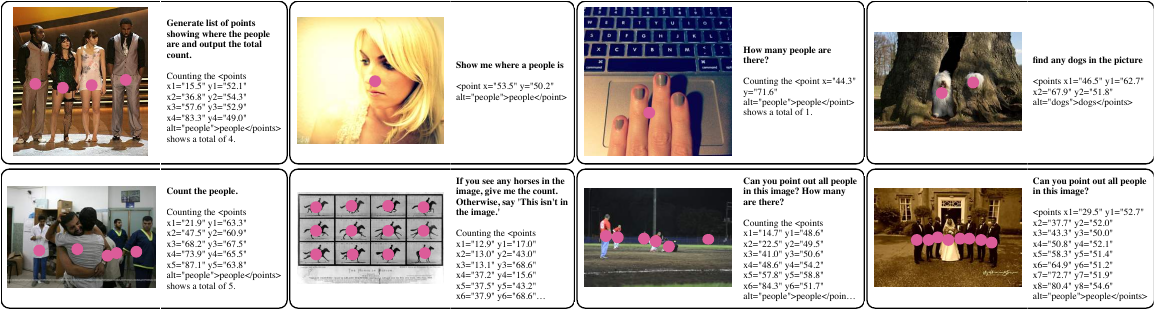}
    \caption{Randomly selected examples from the synthetic \textbf{PixMo-Count} data.}
    \label{fig:qual_count}
\end{figure*}

\begin{figure*}
    \centering
    \includegraphics[width=\linewidth]{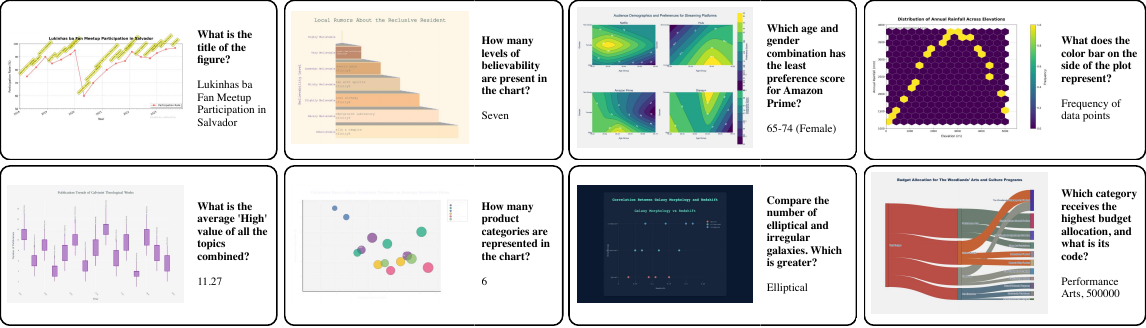}
    \caption{Randomly selected chart examples from the synthetic \textbf{PixMo-Docs} data.}
    \label{fig:qual_docs_charts}
\end{figure*}

\begin{figure*}
    \centering
    \includegraphics[width=\linewidth]{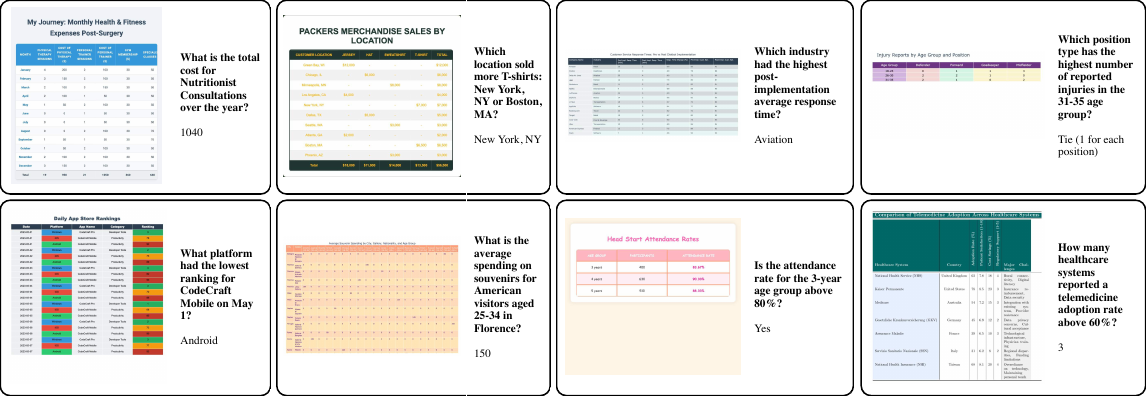}
    \caption{Randomly selected table examples from the synthetic \textbf{PixMo-Docs} data.}
    \label{fig:qual_docs_tables}
\end{figure*}

\begin{figure*}
    \centering
    \includegraphics[width=\linewidth]{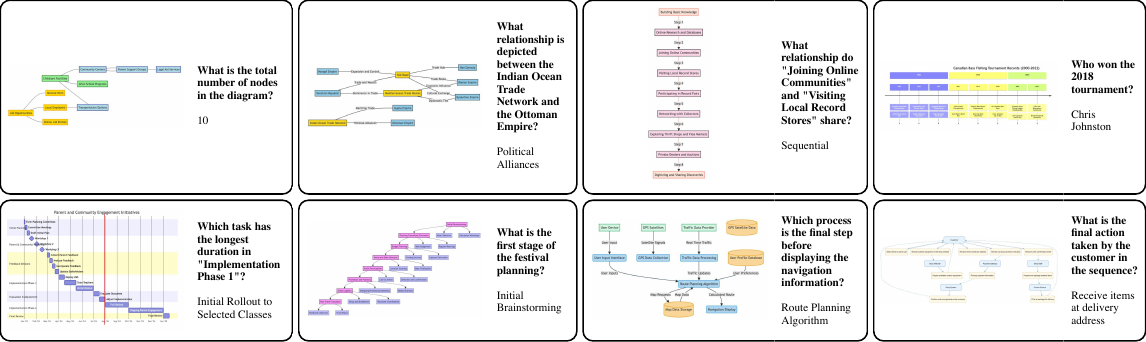}
    \caption{Randomly selected diagram examples from the synthetic \textbf{PixMo-Docs} data.}
    \label{fig:qual_docs_diagrams}
\end{figure*}

\begin{figure*}
    \centering
    \includegraphics[width=\linewidth]{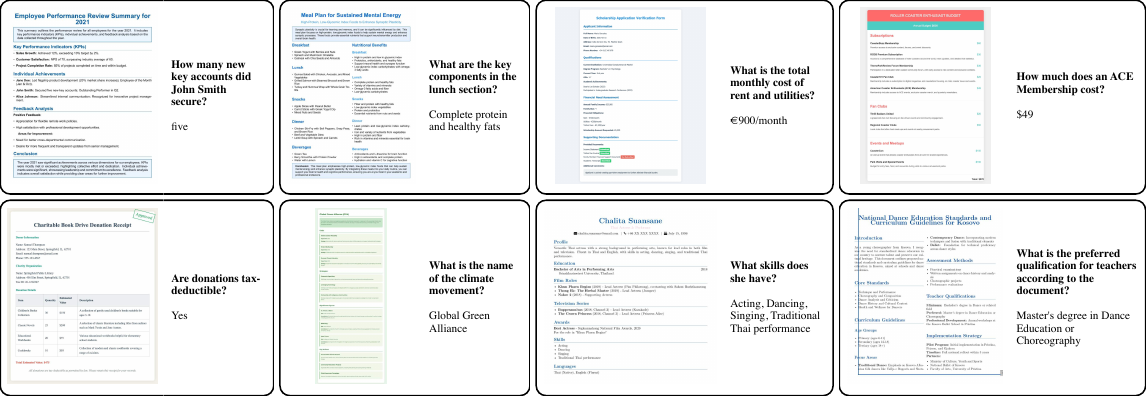}
    \caption{Other randomly selected documents from the synthetic \textbf{PixMo-Docs} data.}
    \label{fig:qual_docs_other}
\end{figure*}

\begin{itemize}
    \item PixMo-Cap - Figure~\ref{fig:qual_cap}
    \item PixMo-AskModelAnything- Figure~\ref{fig:qual_ask_model_anything}
    \item PixMo-Points - Figure~\ref{fig:qual_points}
    \item PixMo-Points with explanations -  Figure~\ref{fig:qual_point_qa}
    \item PixMo-CapQA -  Figure~\ref{fig:qual_cap_qa}
    \item PixMo-Clocks -  Figure~\ref{fig:qual_clocks}
    \item PixMo-Count -  Figure~\ref{fig:qual_count}
    \item PixMo-Docs (charts) -  Figure~\ref{fig:qual_docs_charts}
    \item PixMo-Docs (tables) -  Figure~\ref{fig:qual_docs_tables}
    \item PixMo-Docs (diagrams) -  Figure~\ref{fig:qual_docs_diagrams}
    \item PixMo-Docs (other) -  Figure~\ref{fig:qual_docs_other}
\end{itemize}

\section{Related Work}
\label{supp:related_work}

\paragraph{Vision-language contrastive models.} Vision-language models have become popular in the last few years. Models such as CLIP~\cite{clip} and ALIGN~\cite{jia2021scaling} that are trained on noisy web data provide strong language-aligned image encoders and perform well on downstream classification and image-text retrieval tasks, without any task specific tuning. Previous works~\cite{weston2011wsabie, frome2013devise} proposed similar ideas before transformers~\cite{vaswani2017attention} became popular. Since CLIP was released, other works have focused on making the CLIP pipeline fully open~\cite{xudemystifying, Cherti_2023_CVPR}.  However, vision encoders trained with noisy web data have limitations in discerning details, as discussed in~\cite{tong2024eyes}.

\paragraph{Multimodal LLMs.} Multimodal LLMs often use CLIP-style image encoders and align image embeddings with the LLM input space via a connector module~\cite{liu2023llava, li2024llava, liu2024llavanext, peng2023kosmos2groundingmultimodallarge, Dai2023InstructBLIPTG, wang2024cogvlm, hong2024cogvlm2, Thapa2024DragonflyMZ, McKinzie2024MM1MA, Zhang2024MM15MA}. Some works have also explored using multiple image encoders in tandem with CLIP-style encoders~\cite{tong2024cambrian, lu2024deepseek}, such as using self-supervised learning (SSL) encoders~\cite{oquab2023dinov2}. Many works use a pre-training stage for just the connector weights~\cite{liu2023llava, tong2024cambrian, Cha2024HoneybeeLP, fu2024vita} while others do not have an explicit connector training stage~\cite{karamcheti2024prismatic, beyer2024paligemma, liu2024sphinxx, chen2023pali}. In contrast, two other common architecture strategies are (1) directly connecting the image embedding to different LLM layer embeddings via cross-attention~\cite{alayrac2022flamingo, llama3, Li2023OtterAM, Zhang2024LLaMAAdapterEF} and (2) removing the image encoder and directly inputting the pixels~\cite{fuyu8b, li2023otterhd}. The cross-attention design naturally allows for the introduction of a large number of new parameters, which enables freezing the LLM while still training an effective VLM. This approach has the advantage of maintaining text-only task performance (\cf Table~\ref{table:text-only}).
Due to the compute constraints for training and inference of these models, there has also been a rise in efficient multimodal LLMs~\cite{yuan2024tinygptv, luo2023cheap, chen2024evlmefficientvisionlanguagemodel, lin2024moe, chu2023mobilevlmfaststrong, yao2024minicpm, zhu2024minigpt4}.

The best performing multimodal LLMs~\cite{hurst2024gpt, anthropic2024claude, team2024gemini} are proprietary closed source models. While they are very capable, not much is known about how these models are trained and what data they use.
In contrast, many works release their model weights~\cite{agrawal2024pixtral, wang2024qwen2, abdin2024phi, beyer2024paligemma, llama3} but don't release their training recipes or don't disclose all the data used. Other works provide all the training details and data~\cite{li2024llava, tong2024cambrian, xue2024xgen, laurenccon2024building, zhang2024internlm}, but use data generated by proprietary VLMs such as~\cite{chen2023sharegpt4v}. Hence, there is a need for a fully open SoTA training pipeline that does not use previously trained multimodal LLMs to generate data.

\paragraph{Vision-language instruction tuning datasets.} The rise in popularity of VLMs has also led to a rise of methods to build visual instruction-tuning data. A common approach is to annotate an image with vision models (or use ground-truth annotations), and then use a LLM to generate QA pairs~\cite{liu2023llava, zhu2024minigpt4, Li2023MIMICITMI,you2023ferret} from those annotations. However, these approaches are limited since the automatically generated annotations can be noisy, and even ground-truth annotations often do not comprehensively describe all the details in the image. PixMo-CapQA takes a similar approach but uses the detailed captions from PixMo-Cap which provide more comprehensive image descriptions.  Many recent methods have used proprietary VLMs to annotate images directly~\cite{chen2023sharegpt4v,chen2024allava,liu2024llavanext,wang2023see,liu2023mmc}, which is effective but makes the training pipeline dependent on a closed source VLM.

It is also very common to pair templated instructions with existing annotated datasets to build instruction tuning data (\eg, \cite{Li2023MIMICITMI,lu2024unified,jiang2024mantis,llama3}). 
While Molmo also uses academic datasets, we prefer style tags over natural language instructions since we believe our data, and in particular PixMo-AskModelAnything, provides better training for conversational user interactions.

Our approach to having annotators work with a LLM when generating QA pairs is similar to the approach in~\cite{malaviya2023expertqa}, but we extend this idea to image/language data.

\paragraph{Synthetic vision-language datasets.}
Prior approaches to synthetic chart generation typically only support one or two types of charts~\cite{kafle2018dvqa, methani2020plotqa, kahou2017figureqa}, often with a heavy focus on bar charts or line plots. PixMo-Docs uses code as the text-only representation for the LLM which lets us support much more diverse formats, including heat-maps, violin plots, chord diagrams, geographic plots, tree maps among many others. 
Our use of HTML for document generation is also similar to~\cite{laurenccon2024unlocking}, however we consider many additional approaches to representing documents besides HTML.

Synthetic clock data has been considered~\cite{yang2022s}. Our approach uses real watch faces instead of rendering clocks purely from a simulator, which gives our synthetic data more diversity (\eg, watches with no second hand, stylized decorations or coloring, background images, a separate inner piece to show seconds). Combining these two datasets might yield additional improvements.

\paragraph{VLM grounding.} 
Multimodal LLMs that support grounding language in an image are becoming more common~\cite{Pramanick2023JackOA, Zhang2024Ferretv2AI, peng2023kosmos2groundingmultimodallarge, rasheed2024glamm, xiao2024florence, li2024covlm, Yuan2024OspreyPU}. These works commonly use automated object detectors and/or existing referring expression datasets~\cite{yu2016modeling, GRES,Krishna2016VisualGC} for training data. 
Of these datasets, GRES~\cite{GRES} is most similar to PixMo-Points in that it is human-annotated and includes arbitrary expressions (not just object categories), none-present annotations, and allows expressions to refer to multiple object in the image.
However, it only grounds a limited category of objects (\eg., only COCO categories), and rarely grounds expressions in a large numbers of objects in a single image.
For PixMo-Points we source a diverse set of images and collect points from human annotators. We annotate points instead of segmentation masks, which enables us to collect 1.98M unique referring expression instances with an average of 5.5 points per expression.

\paragraph{Bootstrapping from LLMs.}
Closed text-only LLMs are commonly used for data generation and curation~\cite{liu2023llava}. We also made use of closed text-only LLMs when building several of the PixMo datasets. 
Given our stance against using \emph{VLMs} for building datasets, it's worth justifying the use of closed LLMs. 
It is true using closed LLMs means the current data pipeline is not entirely open.
However, once \emph{open} LLMs (\eg,~\cite{OLMo}) become sufficiently good they can be used in place of closed ones to build a dataset functionally equivalent to PixMo. Our philosophy is that we should not wait for open LLM research to achieve this goal and instead we should pursue research on building open VLMs \emph{in parallel}. We note that using one VLM to build another VLM is entirely different than using an LLM, because the dependency is circular and therefore cannot result in a fully open system at a later point in time.

\clearpage
\label{supp:qualitative}

\clearpage
{
    \footnotesize
    \bibliographystyle{ieeenat_fullname}
    \bibliography{mmolmo}
}

\end{document}